\pdfoutput=1

\documentclass[10pt,twocolumn,letterpaper]{article}

\usepackage{cvpr}
\usepackage{times}
\usepackage{epsfig}
\usepackage{graphicx}
\usepackage{amsmath}
\usepackage{amssymb}

\usepackage{amsmath,amsthm,verbatim,amssymb,amsfonts,amscd, graphicx}
\usepackage{graphics}
\usepackage[lined, boxruled, linesnumberedhidden]{algorithm2e}
\usepackage[T1]{fontenc} 
\usepackage{fancyhdr}

\usepackage{graphicx}
\graphicspath{{./figures/}}

\usepackage{xspace}
\makeatletter
\DeclareRobustCommand\onedot{\futurelet\@let@token\@onedot}
\def\@onedot{\ifx\@let@token.\else.\null\fi\xspace}

\def\eg{\emph{e.g}\onedot} 
\def\ie{\emph{i.e}\onedot}

\def\etal{\emph{et al}\onedot}
\makeatother

{
\fancyhf{}
\lhead{\small This paper has been submitted for publication on November 15, 2016.}
}

\cvprfinalcopy 

\def\cvprPaperID{767} 

\ifcvprfinal\fi
\begin{document}

\title{Learning from Simulated and Unsupervised Images through Adversarial Training }

\author{Ashish Shrivastava, Tomas Pfister, Oncel Tuzel, Josh Susskind, Wenda Wang, Russ Webb \\ 
Apple Inc\\
{\tt\small \{a\_shrivastava,tpf,otuzel,jsusskind,wenda\_wang,rwebb\}@apple.com}
}

\maketitle

\thispagestyle{fancy}
\begin{abstract}
With recent progress in graphics, it has become more tractable to train models on synthetic images, potentially avoiding the need for expensive annotations. 
However, learning from synthetic images may not achieve the desired performance due to a gap between synthetic and real image distributions. 
To reduce this gap, we propose Simulated+Unsupervised (S+U) learning, where the task is to learn a model to improve the realism of a simulator's output using unlabeled real data, while preserving the annotation information from the simulator.
We develop a method for S+U learning that uses an adversarial network similar to Generative Adversarial Networks (GANs), but with synthetic images as inputs instead of random vectors. 
We make several key modifications to the standard GAN algorithm to preserve annotations, avoid artifacts, and stabilize training: (i)~a `self-regularization' term, (ii)~a local adversarial loss, and (iii)~updating the discriminator using a history of refined images. 
We show that this enables generation of highly realistic images, which we demonstrate both qualitatively and with a user study.
We quantitatively evaluate the generated images by training models for gaze estimation and hand pose estimation.
We show a significant improvement over using synthetic images, and achieve state-of-the-art results on the MPIIGaze dataset without any labeled real data.
\end{abstract}

\begin{figure}
\centering
\includegraphics[width=\linewidth]{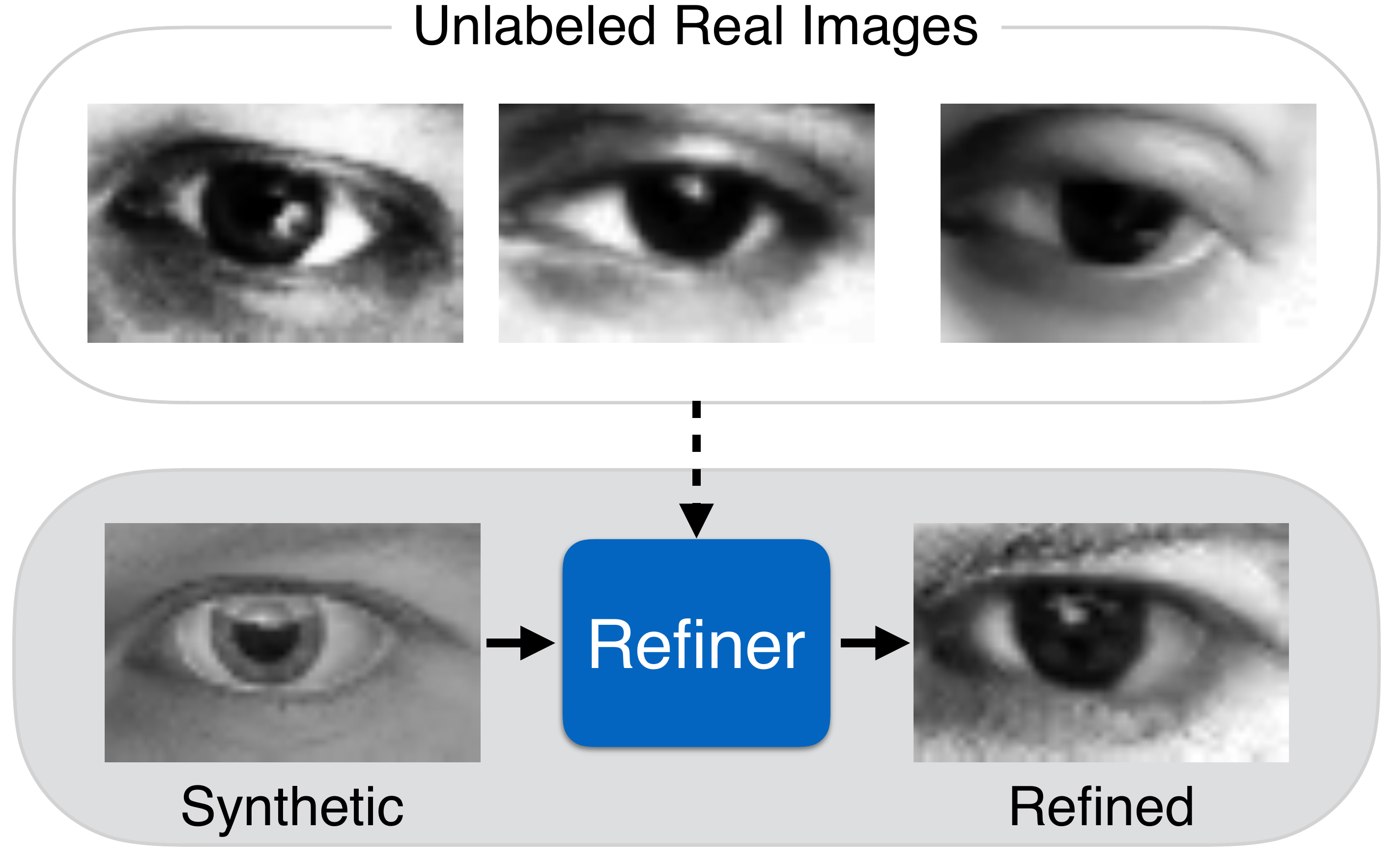}
\caption{Simulated+Unsupervised (S+U) learning. The task is to learn a model that improves the realism of synthetic images from a simulator using unlabeled real data, while preserving the annotation information.}
\label{fig:intro}
\end{figure}

\section{Introduction}

Large labeled training datasets are becoming increasingly important with the recent rise in high capacity deep neural networks~\cite{imagenet_cvpr09, mscoco, ZhangLL16, ZhangLL16, youtub8M_16, Nagaraja2016, openimages}.
However, labeling such large datasets is expensive and time-consuming.
Thus, the idea of training on synthetic instead of real images has become appealing because the annotations are automatically available. 
Human pose estimation with Kinect~\cite{Shotton13} and, more recently, a plethora of other tasks have been tackled using synthetic data~\cite{Wood16,Wang15,qiu2016unrealcv,Shafaei16}.
However, learning from synthetic images can be problematic due to a gap between synthetic and real image distributions -- synthetic data is often not realistic enough, leading the network to learn details only present in synthetic images and failing to generalize well on real images.
One solution to closing this gap is to improve the simulator.
However, increasing the realism is often computationally expensive, the content modeling takes a lot of hard work, and even the best rendering algorithms may still fail to model all the characteristics of real images. 
This lack of realism may cause models to overfit to `unrealistic' details in the synthetic images.

In this paper, we propose Simulated+Unsupervised (S+U) learning, where the goal is to improve the realism of synthetic images from a simulator using unlabeled real data.
The improved realism enables the training of better machine learning models on large datasets without any data collection or human annotation effort. 
In addition to adding realism, S+U learning should preserve annotation information for training of machine learning models -- \eg the gaze direction in Figure~\ref{fig:intro} should be preserved.
Moreover, since machine learning models can be sensitive to artifacts in the synthetic data, S+U learning should generate images without artifacts.

\begin{figure}
\centering
\includegraphics[width=\linewidth]{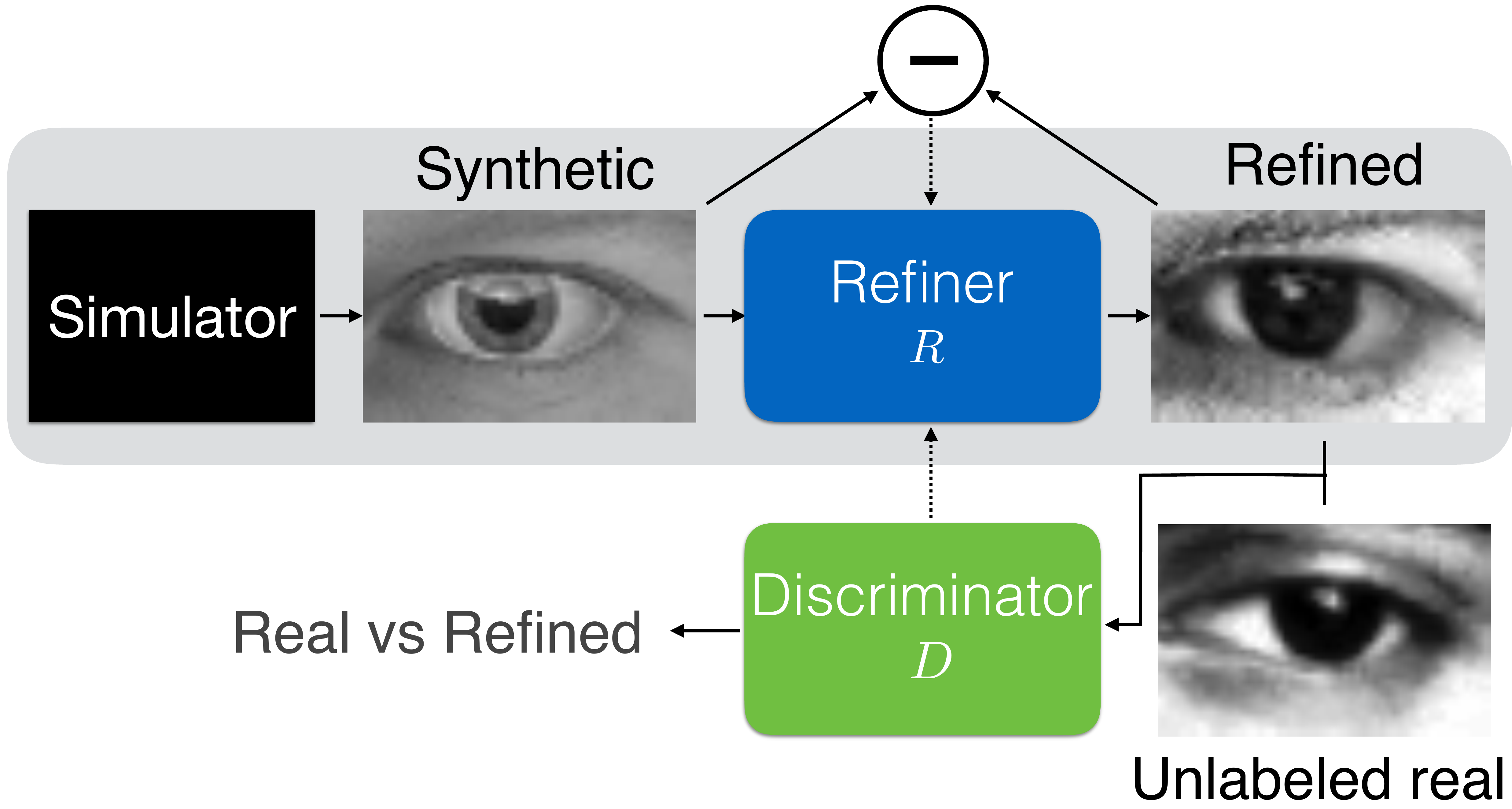}
\caption{Overview of SimGAN. 
We refine the output of the simulator with a refiner neural network, $R$, that minimizes the combination of a local adversarial loss and a `self-regularization' term. 
The adversarial loss `fools' a discriminator network, $D$, that classifies an image as real or refined. 
The self-regularization term minimizes the image difference between the synthetic and the refined images. 
The refiner network and the discriminator network are updated alternately.}
\label{fig:method_overview}
\end{figure}

We develop a method for S+U learning, which we term SimGAN, that refines synthetic images from a simulator using a neural network which we call the `refiner network'.
Figure~\ref{fig:method_overview} gives an overview of our method: a synthetic image is generated with a black box simulator and is refined using the refiner network. 
To add realism, we train our refiner network using an adversarial loss, similar to Generative Adversarial Networks (GANs)~\cite{Goodfellow14}, such that the refined images are indistinguishable from real ones using a discriminative network.
To preserve the annotations of synthetic images, we complement the adversarial loss with a self-regularization loss that penalizes large changes between the synthetic and refined images.
Moreover, we propose to use a fully convolutional neural network that operates on a pixel level and preserves the global structure, rather than holistically modifying the image content as in \eg a fully connected encoder network.
The GAN framework requires training two neural networks with competing goals, which is known to be unstable and tends to introduce artifacts~\cite{Salimans16}.
To avoid drifting and introducing spurious artifacts while attempting to fool a single stronger discriminator, we limit the discriminator's receptive field to local regions instead of the whole image, resulting in multiple local adversarial losses per image. 
Moreover, we introduce a method for improving the stability of training by updating the discriminator using a history of refined images rather than only the ones from the current refiner network. 
\paragraph{Contributions:}
\begin{enumerate}
\item We propose S+U learning that uses unlabeled real data to refine the synthetic images. 
\item We train a refiner network to add realism to synthetic images using a combination of an adversarial loss and a self-regularization loss.
\item We make several key modifications to the GAN training framework to stabilize training and prevent the refiner network from producing artifacts.
\item We present qualitative, quantitative, and user study experiments showing that the proposed framework significantly improves the realism of the simulator output.
We achieve state-of-the-art results, without any human annotation effort, by training deep neural networks on the refined output images.
\end{enumerate}

\subsection{Related Work}
The GAN framework learns two networks (a generator and a discriminator) with competing losses. 
The goal of the generator network is to map a random vector to a realistic image, whereas the goal of the discriminator is to distinguish the generated from the real images.
The GAN framework was first introduced by Goodfellow~\etal~\cite{Goodfellow14} to generate visually realistic images and, since then, many improvements and interesting applications have been proposed~\cite{Salimans16}.
Wang and Gupta~\cite{Wang2016} use a Structured GAN to learn surface normals and then combine it with a Style GAN to generate natural indoor scenes. 
Im~\etal~\cite{Im2015} propose a recurrent generative model trained using adversarial training. 
The recently proposed iGAN~\cite{zhu2016generative} enables users to change the image interactively on a natural image manifold.  
CoGAN by Liu~\etal~\cite{liu2016coupled} uses coupled GANs to learn a joint distribution over images from multiple modalities without requiring tuples of corresponding images, achieving this by a weight-sharing constraint that favors the joint distribution solution.  
Chen~\etal~\cite{Chen16} propose InfoGAN, an information-theoretic extension of GAN, that allows learning of meaningful representations. 
Tuzel~\etal~\cite{Tuzel16} tackled image super-resolution for face images with GANs.  
Li and Wand~\cite{Li2016} propose a Markovian GAN for efficient texture synthesis. 
Lotter~\etal~\cite{Lotter15} use adversarial loss in an LSTM network for visual sequence prediction.
Yu~\etal~\cite{SeqGan_Yu16} propose the SeqGAN framework that uses GANs for reinforcement learning. 
Yoo~\etal~\cite{Yoo16} tackle pixel-level semantic transfer learning with GANs.
Style transfer~\cite{Gatys16} is also closely related to our work.
Many recent works have explored related problems in the domain of generative models, such as PixelRNN~\cite{vandenOord16} that predicts pixels sequentially with an RNN with a softmax loss.
The generative networks focus on generating images using a random noise vector; thus, in contrast to our method, the generated images do not have any annotation information that can be used for training a machine learning model.

Many efforts have explored using synthetic data for various prediction tasks, including gaze estimation~\cite{Wood16}, text detection and classification in RGB images~\cite{Gupta16,Jaderberg16}, font recognition~\cite{Wang15}, object detection~\cite{Gupta14,Peng15}, hand pose estimation in depth images~\cite{tompson14NYU, Supancic15}, scene recognition in RGB-D~\cite{Handa16}, semantic segmentation of urban scenes~\cite{Ros_2016_CVPR}, and human pose estimation~\cite{Park15,Darrell15,LeCun04a,Ionescu14,Pishchulin12,Rogez16}. 
Gaidon~\etal~\cite{Gaidon16} show that pre-training a deep neural network on synthetic data leads to improved performance.
Our work is complementary to these approaches, where we improve the realism of the simulator using unlabeled real data.

Ganin and Lempitsky~\cite{Ganin14} use synthetic data in a domain adaptation setting where the learned features are invariant to the domain shift between synthetic and real images. 
Wang~\etal~\cite{Wang15} train a Stacked Convolutional Auto-Encoder on synthetic and real data to learn the lower-level representations of their font detector ConvNet.
Zhang~\etal~\cite{Zhang15} learn a Multichannel Autoencoder to reduce the domain shift between real and synthetic data.
In contrast to classical domain adaptation methods that adapt the features with respect to a specific prediction task, we bridge the gap between image distributions through adversarial training. 
This approach allows us to generate realistic training images which can be used to train any machine learning model, potentially for multiple tasks.

Johnson~\etal~\cite{johnson11cg2real} transfer the style from a set of real images to  the synthetic image by co-segmenting and then identifying similar regions. 
This approach requires users to select the top few matches from an image database. 
In contrast, we propose an end-to-end solution that does not require user intervention at inference time.

\section{S+U Learning with SimGAN}

The goal of Simulated+Unsupervised learning is to use a set of unlabeled real images $\mathbf y_i \in \mathcal Y$ to learn a refiner $R_{\boldsymbol{\theta}} (\mathbf x)$ that refines a synthetic image $\mathbf x $, where $\boldsymbol {\theta} $  are the function parameters. 
Let the refined image be denoted by  $\tilde{\mathbf x}$, then
$\tilde{\mathbf x} := R_{\boldsymbol{\theta}}(\mathbf x).$
The key requirement for S+U learning is that the refined image $\tilde{\mathbf x} $ should look like a real image in appearance while preserving the annotation information from the simulator.

To this end, we propose to learn $\boldsymbol \theta$ by minimizing a combination of two losses:
\begin{align}
{\mathcal L}_R(\boldsymbol \theta) = \sum_i  \ell_{\text{real}} (\boldsymbol \theta; \mathbf x_i, \mathcal Y ) + \lambda \ell_{\text{reg}} (\boldsymbol \theta;  {\mathbf x_i}),
\label{eq:L_R_self_reg}
\end{align}
where $\mathbf x_i$ is the $i^{\text th}$ synthetic training image. 
The first part of the cost, $ \ell_{\text{real}}$, adds realism to the synthetic images, while the second part, $\ell_{\text{reg}}$, preserves the annotation information.
In the following sections, we expand this formulation and provide an algorithm to optimize for $\boldsymbol \theta$.

\subsection{Adversarial Loss with Self-Regularization}

To add realism to the synthetic image, we need to bridge the gap between the distributions of synthetic and real images. 
An ideal refiner will make it impossible to classify a given image as real or refined with high confidence. 
This need motivates the use of an adversarial discriminator network, $D_{\boldsymbol{\phi}} $, that is trained to classify images as real vs refined, where $\boldsymbol \phi $ are the parameters of the discriminator network. 
The adversarial loss used in training the refiner network, $R$, is responsible for `fooling' the network $D$ into classifying the refined images as real. 
Following the GAN approach~\cite{Goodfellow14}, we model this as a two-player minimax game, and update the refiner network, $R_{\boldsymbol{\theta}}$, and the discriminator network, $D_{\boldsymbol{\phi}}$, alternately. 
Next, we describe this intuition more precisely.

The discriminator network updates its parameters by minimizing the following loss:
\begin{align}
{\mathcal L}_D (\boldsymbol \phi) =- \sum_i\log(D_{\boldsymbol \phi}(\tilde{\mathbf x}_i)) - \sum_j\log (1 - D_{\boldsymbol \phi}(\mathbf y_j)) .
\label{eq:loss_D}
\end{align}
This is equivalent to cross-entropy error for a two class classification problem where $D_{\boldsymbol \phi}(.)$ is the probability of the input being a synthetic image, and $1- D_{\boldsymbol \phi}(.)$  that of a real one. 
We implement $D_{\boldsymbol \phi}$ as a ConvNet whose last layer outputs the probability of the sample being a refined image. 
For training this network, each mini-batch consists of randomly sampled refined synthetic images $\tilde{\mathbf x}_i$'s and real images $\mathbf y_j$'s. 
The target labels for the cross-entropy loss layer are $0$ for every $\mathbf y_j$, and $1$ for every $\tilde{\mathbf x}_i$. 
Then $\boldsymbol \phi$ for a mini-batch is updated by taking a stochastic gradient descent (SGD) step on the  mini-batch loss gradient. 

In our implementation, the realism loss function  $\ell_{\text{real}}$ in ~\eqref{eq:L_R_self_reg} uses the trained discriminator $D$ as follows:
\begin{align}
\ell_{\text{real}}  (\boldsymbol \theta; \mathbf x_i, \mathcal Y ) = -  \log (1 - D_{\boldsymbol \phi}(R_{\boldsymbol \theta}(\mathbf x_i) ) ) .
\end{align}
By minimizing this loss function, the refiner forces the discriminator to fail classifying the refined images as synthetic.
In addition to generating realistic images, the refiner network should preserve the annotation information of the simulator. 
For example, for gaze estimation the learned transformation should not change the gaze direction, and for hand pose estimation the location of the joints should not change. 
This restriction is an essential ingredient to enable training a machine learning model that uses the refined images with the simulator's annotations. 
For this purpose, we propose using a self-regularization loss that minimizes per-pixel difference between a feature transform of the synthetic and refined images,  $\ell_{\text{reg}} = \|\psi(\tilde{\mathbf x}) - \mathbf x\|_1 $, where
$\psi$ is the mapping from image space to a feature space, and $\|.\|_1$ is the L1 norm.
The feature transform can be an identity map ($\psi(\mathbf x) = \mathbf x$), image derivatives, mean of color channels, or a learned transformation such as a convolutional neural network. 
In this paper, unless otherwise stated, we used the identity map as the feature transform. 
Thus, the overall refiner loss function \eqref{eq:L_R_self_reg} used in our implementation is:
\begin{align}
{\mathcal L}_R (\boldsymbol \theta) = - \sum_i \log (1 - D_{\boldsymbol \phi}(R_{\boldsymbol \theta}(\mathbf x_i) ) )   \nonumber \\ 
	 + \lambda \| \psi(R_{\boldsymbol \theta}(\mathbf x_i)) - \psi(\mathbf x_i)\|_1.
\label{eq:L_R_i_self_reg}
\end{align}
We implement $R_{\boldsymbol \theta} $ as a fully convolutional neural net without striding or pooling,
modifying the synthetic  image on a pixel level, rather than holistically modifying the image content as in \eg a fully connected encoder network, thus preserving the global structure and annotations.
We learn  the refiner and discriminator parameters by minimizing $ {\mathcal L}_R (\boldsymbol \theta)$ and ${\mathcal L}_D (\boldsymbol \phi)$ alternately. 
While updating the parameters of $R_{\boldsymbol \theta}$, we keep $\boldsymbol \phi$ fixed, and while updating $D_{\boldsymbol \phi}$,  we fix  $\boldsymbol \theta$. We summarize this training procedure in Algorithm~\ref{alg:training_GAN_w_L1}.

\begin{algorithm}[t]
\KwIn {Sets of synthetic  images $\mathbf x_i \in \mathcal X$ , and real images  $ \mathbf y_j \in \mathcal Y$, max number of steps ($T$), number of discriminator network updates per step ($K_d$), number of generative network updates per step ($K_g$).}
\KwOut {ConvNet model $R_{\boldsymbol \theta}$.}
\For {$t = 1, \dots, T$ } 
{
\For {$k = 1, \dots, K_g$ } 
{
1. Sample a mini-batch of  synthetic images $\mathbf x_i$. \\
2. Update $\boldsymbol \theta$ by taking a SGD step on mini-batch loss $ {\mathcal L}_R(\boldsymbol \theta)$ in ~\eqref{eq:L_R_i_self_reg} .
}
\For {$k = 1, \dots, K_d$ } 
{
1. Sample a mini-batch of  synthetic images $ \mathbf x_i$, and real images $\mathbf y_j$. \\
2. Compute $\tilde{\mathbf x}_i = R_{\boldsymbol \theta}(\mathbf x_i)$ with current $\boldsymbol \theta$. \\
3. Update $\boldsymbol \phi$ by taking a SGD step  on mini-batch loss $  {\mathcal L}_D(\boldsymbol \phi)$ in ~\eqref{eq:loss_D}.
}

}
\caption{Adversarial training of refiner network $R_{\boldsymbol \theta}$}
\label{alg:training_GAN_w_L1}
\vspace{-0.2cm}
\end{algorithm}

\subsection{Local Adversarial Loss}
Another key requirement for the refiner network is that it should learn to model the real image characteristics without introducing any artifacts. When we train a single strong discriminator network, the refiner network tends to over-emphasize certain image features to fool the current discriminator network, leading to drifting and producing artifacts.
A key observation is that any local patch sampled from the refined image should have similar statistics to a real image patch. 
Therefore, rather than defining a global discriminator network, we can define a discriminator network that classifies all local image patches separately.
This division not only limits the receptive field, and hence the capacity of the discriminator network, but also provides many samples per image for learning the discriminator network. 
The refiner network is also improved by having multiple `realism loss' values per image. 

\begin{figure}[t]
\vspace{-0.2cm}
\centering
\includegraphics[width=1.0\linewidth]{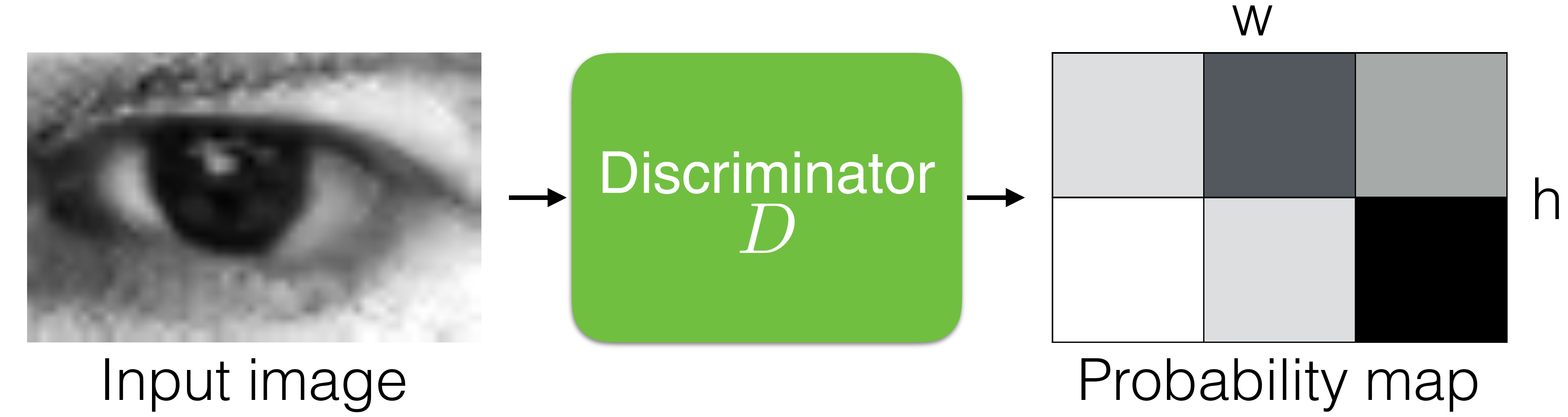} \\
\caption{Illustration of local adversarial loss. 
The discriminator network outputs a $w \times h$ probability map.
The adversarial loss function is the sum of the cross-entropy losses over the local patches.  
}
\label{fig:local_adversarial_loss_illust}
\end{figure}

In our implementation, we design the discriminator $D$ to be a fully convolutional network that outputs $w \times h $ dimensional probability map of patches belonging to the fake class, where $w \times h$ are the number of local patches in the image. While training the refiner network, we sum the cross-entropy loss values over  $w \times h$ local patches, as illustrated in Figure~\ref{fig:local_adversarial_loss_illust}. 

\begin{figure}[t]
\centering
\includegraphics[width=0.8\linewidth]{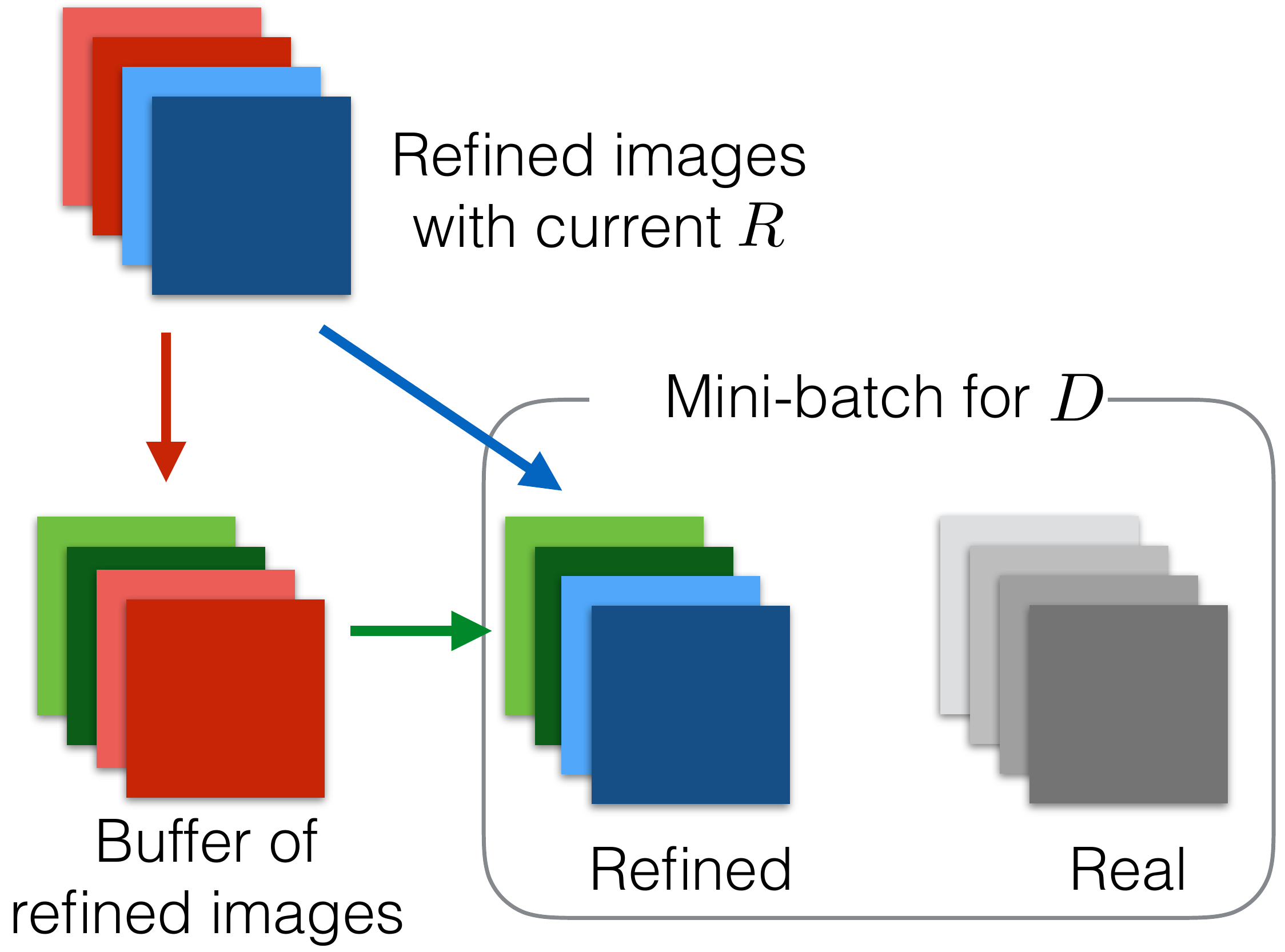} \\
\caption{Illustration of using a history of refined images. See text for details.
}
\label{fig:history_illust}
\end{figure}

\begin{figure*}[t]
\centering
\includegraphics[width=\linewidth]{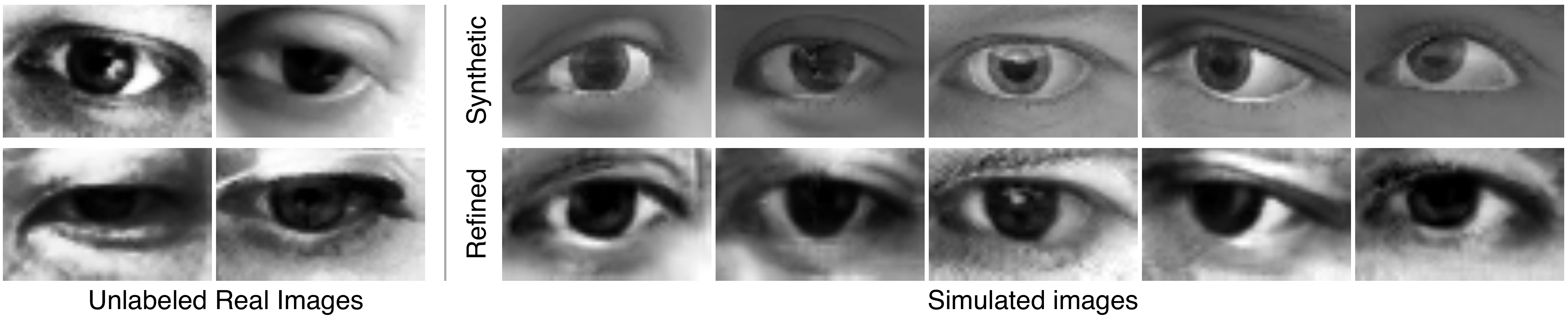} 
\caption{Example output of SimGAN for the UnityEyes gaze estimation dataset~\cite{Wood16}. 
(Left) real images from MPIIGaze~\cite{Zhang15a}. 
Our refiner network does not use any label information from MPIIGaze dataset at training time. 
(Right) refinement results on UnityEye. 
The skin texture and the iris region in the refined synthetic images are qualitatively significantly more similar to the real images than to the synthetic images.
More examples are included in the supplementary material.
}
\label{fig:results_qualitative_gaze}
\vspace{-0.2cm}
\end{figure*}

\subsection{Updating Discriminator using a History of Refined Images}
Another problem of adversarial training is that the discriminator network only focuses on the latest refined images. 
This lack of memory may cause (i)~divergence of the adversarial training, and (ii)~the refiner network re-introducing the artifacts that the discriminator has forgotten about. 
Any refined image generated by the refiner network at any time during the entire training procedure is a `fake' image for the discriminator.
Hence, the discriminator should be able to classify all these images as fake.
Based on this observation, we introduce a method to improve the stability of adversarial training by updating the discriminator using a history of refined images, rather than only the ones in the current mini-batch.
We slightly modify Algorithm~\ref{alg:training_GAN_w_L1} to have a buffer of refined images generated by previous networks. 
Let $B$ be the size of the buffer and $b$ be the mini-batch size used in Algorithm~\ref{alg:training_GAN_w_L1}. 
At each iteration of discriminator training, we compute the discriminator loss function by sampling  $b/2$ images from the current refiner network, and sampling an additional $b/2$ images from the buffer to update parameters $\boldsymbol \phi$. 
We keep the size of the buffer, $B$, fixed. 
After each training iteration, we randomly replace $b/2$ samples in the buffer with the newly generated refined images. 
This procedure is illustrated in Figure~\ref{fig:history_illust}.

In contrast to our approach, Salimans \etal~\cite{Salimans16} used  a running average of the model parameters to stabilize the training. 
Note that these two approaches are complementary and can be used together.

\section{Experiments}

We evaluate our method for appearance-based gaze estimation in the wild on the MPIIGaze dataset~\cite{Wood16,Zhang15a}, and hand pose estimation on the NYU hand pose dataset of depth images~\cite{tompson14NYU}. 
We use a fully convolutional refiner network with ResNet blocks for all of our experiments.

\subsection{Appearance-based Gaze Estimation}

Gaze estimation is a key ingredient for many human computer interaction (HCI) tasks. 
However, estimating the gaze direction from an eye image is challenging, especially when the image is of low quality, \eg from a laptop or a mobile phone camera -- annotating the eye images with a gaze direction vector is challenging even for humans.
Therefore, to generate large amounts of annotated data, several recent approaches~\cite{Wood16,Zhang15a} train their models on large amounts of synthetic data. 
Here, we show that training with the refined synthetic images generated by SimGAN significantly outperforms the state-of-the-art for this task.

The gaze estimation dataset consists of 1.2M synthetic images from the UnityEyes simulator~\cite{Wood16}  and 214K real images from the MPIIGaze dataset~\cite{Zhang15a} -- samples shown in Figure~\ref{fig:results_qualitative_gaze}.
MPIIGaze is a very challenging eye gaze estimation dataset captured under extreme illumination conditions.
For UnityEyes, we use a single generic rendering environment to generate training data without any dataset-specific targeting.

\vspace{-0.15in}
\paragraph{Qualitative Results :}
Figure~\ref{fig:results_qualitative_gaze} shows examples of synthetic, real and refined images from the eye gaze dataset.
As shown, we observe a significant qualitative improvement of the synthetic images: SimGAN successfully captures the skin texture, sensor noise and the appearance of the iris region in the real images. Note that our method preserves the annotation information (gaze direction) while improving the realism.

\begin{figure}
\centering
\newcommand\expimagewidth{0.32}
\begin{tabular}{ccc}
\hspace{-0.2cm}\includegraphics[width=\expimagewidth\linewidth]{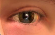} & \hspace{-0.4cm} \includegraphics[width=\expimagewidth\linewidth]{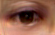} &\hspace{-0.4cm} \includegraphics[width=\expimagewidth\linewidth]{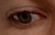} \\
\small Synthetic &   \small Refined  & \small Sample real  \\
\end{tabular}
\caption{Self-regularization in feature space for color images.
}
\label{fig:gaze_l1_on_Y}
\vspace{-0.4cm}
\end{figure}

\paragraph{Self-regularization in Feature Space:}
When the synthetic and real images have significant shift in the distribution, a pixel-wise L1 difference may be restrictive. In such cases, we can replace the identity map with an alternative feature transform. For example, in Figure~\ref{fig:gaze_l1_on_Y}, we use the mean of RGB channels for color image refinement. As shown, the network trained using this feature transform is able to generate realistic color images.
Note that in our quantitative experiments we still use grayscale images because gaze estimation is better tackled in grayscale due to added invariance~\cite{Wood16, Zhang15a}. 

\paragraph{`Visual Turing Test':}
To quantitatively evaluate the visual quality of the refined images, we designed a simple user study where subjects were asked to classify images as real or refined synthetic. 
Each subject was shown a random selection of $50$ real images and $50$ refined images in a random order, and was asked to label the images as either real or refined. 
The subjects were constantly shown $20$ examples of real and refined images while performing the task. 
The subjects found it very hard to tell the difference between the real images and the refined images. 
In our aggregate analysis, $10$ subjects chose the correct label $517$ times out of $1000$ trials ($p=0.148$), meaning they were not able to reliably distinguish real images from synthetic.
Table~\ref{tab:results_gaze_turing} shows the confusion matrix. 
In contrast, when testing on original synthetic images vs real images, we showed $10$ real and $10$ synthetic images per subject, and the subjects chose correctly $162$ times out of $200$ trials ($p \le 10^{-8}$), which is significantly better than chance.

\begin{table}[t]
\centering
\begin{tabular}{|c|c|c|} \hline
&Selected as real & Selected as synt \\ \hline
Ground truth real & 224 & 276 \\ \hline
Ground truth synt & 207 & 293 \\ \hline
\end{tabular}
\caption{Results of the `Visual Turing test' user study for classifying  real vs  refined images.
The average human classification accuracy was $51.7\%$ (chance = $50\%$). 
}
\label{tab:results_gaze_turing}
\end{table}

\begin{figure}
\centering
\includegraphics[width=\linewidth]{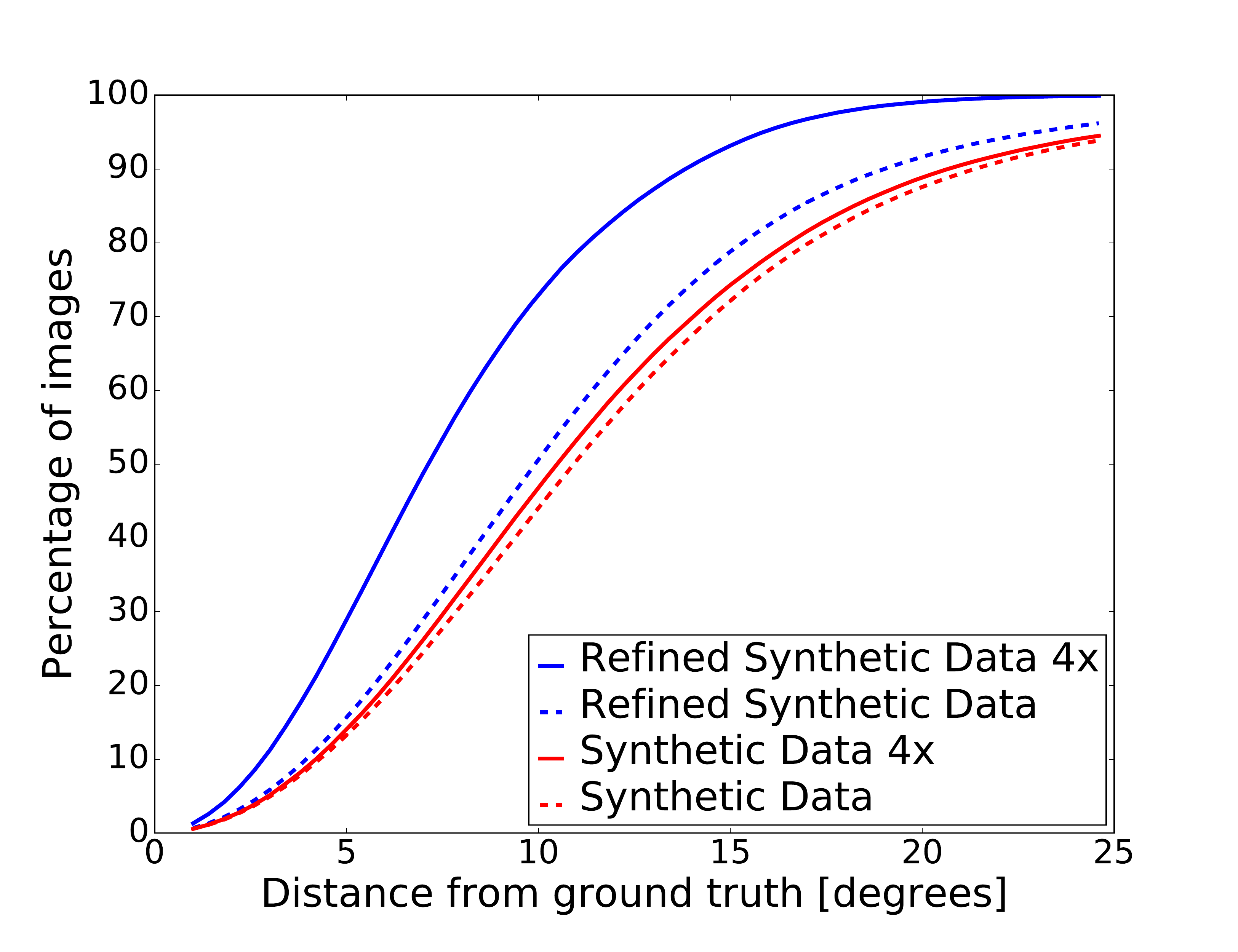} 
\caption{Quantitative results for appearance-based gaze estimation on the MPIIGaze dataset with real eye images.  
The plot shows cumulative curves as a function of degree error as compared to the ground truth eye gaze direction, for different numbers of training examples of data.
}
\label{fig:results_plots_gaze}
\end{figure}

\vspace{-0.2in}
\paragraph{Quantitative Results:}
We train a simple convolutional neural network (CNN) similar to~\cite{Zhang15a} to predict the eye gaze direction (encoded by a 3-dimensional vector for $x,y,z$) with $l_2$ loss. We train on UnityEyes and test on MPIIGaze.
Figure~\ref{fig:results_plots_gaze} and Table~\ref{tab:results_table_gaze} compare the performance of a gaze estimation CNN trained on synthetic data to that of another CNN trained on refined synthetic data, the output of SimGAN. 
We observe a large improvement in performance from training on the SimGAN output, a $22.3\%$ absolute percentage improvement. 
We also observe a large improvement by using more training data -- here 4x refers to $100\%$ of the training dataset.
The quantitative evaluation confirms the value of the qualitative improvements observed in Figure~\ref{fig:results_qualitative_gaze}, and shows that machine learning models generalize significantly better using SimGAN.

Table~\ref{tab:MPIIGaze_result} shows a comparison to the state-of-the-art.
Training the CNN on the refined images outperforms the state-of-the-art on the MPIIGaze dataset, with a relative improvement of $21\%$.
This large improvement shows the practical value of our method in many HCI tasks.

\begin{table}[]
\centering
\begin{tabular}{|l|c|} \hline
Training data  & \% of images within $d$  \\ \hline 
Synthetic Data & 62.3 \\
Synthetic Data 4x & 64.9 \\
Refined Synthetic Data & 69.4 \\
Refined Synthetic Data 4x & {\bf 87.2} \\
   \hline
\end{tabular}
\caption{Comparison of a gaze estimator trained on synthetic data and the output of SimGAN.
The results are at distance $d=7$ degrees from ground truth.
Training on the output of SimGAN outperforms training on synthetic data by $22.3\%$.
}
\label{tab:results_table_gaze}
\end{table}

\begin{table}[]
\centering
\begin{tabular}{|l|c|c|} \hline
 Method  & R/S & Error   \\ \hline 
  Support Vector Regression (SVR)~\cite{Schneider2014} & R &  $16.5$ \\
  Adaptive Linear Regression ALR)~\cite{Lu2014} & R & $16.4$ \\
  Random Forest  (RF)~\cite{Sugano2014} & R & $15.4$ \\
  kNN with UT Multiview~\cite{Zhang15a} & R & $16.2$ \\
 CNN with UT Multiview~\cite{Zhang15a}& R & $13.9$ \\ 
  k-NN with UnityEyes~\cite{Wood16}& S & $9.9$ \\
 CNN with UnityEyes Synthetic Images & S & 11.2 \\ 
 CNN with UnityEyes Refined Images & S & {\bf 7.8} \\ \hline
 
\end{tabular}
\caption{Comparison of SimGAN to the state-of-the-art on the MPIIGaze dataset of real eyes.
The second column indicates whether the methods are trained on Real/Synthetic data.
The error the is mean eye gaze estimation error in degrees.
Training on refined images results in a $2.1$ degree improvement, a relative $21\%$ improvement compared to the state-of-the-art.}
\label{tab:MPIIGaze_result}
\end{table}

\paragraph{Preserving Ground Truth:}
To quantify that the ground truth gaze direction doesn't change significantly, we manually labeled the ground truth pupil centers in $100$ synthetic and refined images by fitting an ellipse to the pupil. 
This is an approximation of the gaze direction, which is difficult for humans to label accurately.
The absolute difference between the estimated pupil center of synthetic and corresponding refined image is quite small: $1.1\pm0.8$px (eye width=$55$px).

\begin{figure*}[t]
\centering
\includegraphics[width=1.0\linewidth]{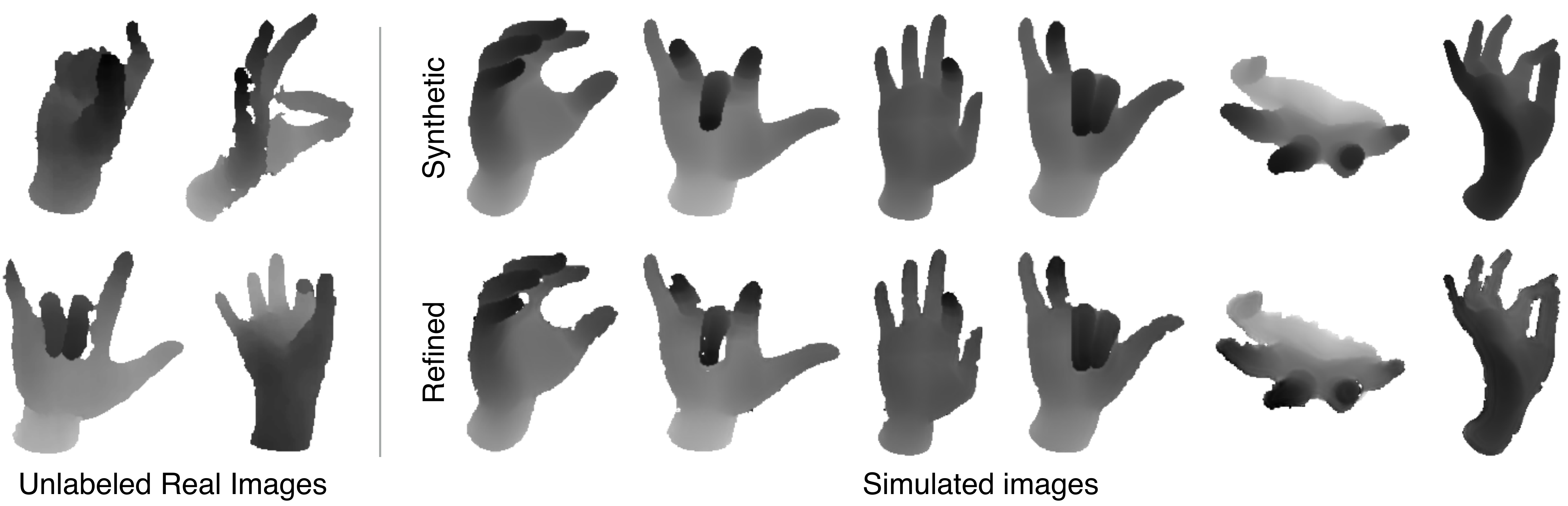} 
\caption{Example refined test images for the NYU hand pose dataset~\cite{tompson14NYU}. 
(Left) real images, (right) synthetic images and the corresponding refined output images from the refiner network. 
The major source of noise in the real images is the non-smooth depth boundaries that the refiner networks learns to model. 
}
\vspace{-0.2cm}
\label{fig:results_qualitative_hand}
\end{figure*}

\paragraph{Implementation Details:} 
The refiner network, $R_{\boldsymbol \theta}$, is a residual network (ResNet)~\cite{He2015}. 
Each ResNet block consists of two convolutional layers containing $64$ feature maps. 
An input image of size $55 \times 35$ is convolved with $3 \times 3$  filters that output $64$ feature maps. 
The output is passed through $4$ ResNet blocks. The output of the last ResNet block is passed to a $1 \times 1$ convolutional layer producing $1$ feature map corresponding to the refined synthetic image.

The discriminator network, $D_{\boldsymbol \phi}$, contains $5$ convolution layers and $2$ max-pooling layers as follows: 
(1)~Conv3x3, stride=2, feature maps=96, 
(2)~Conv3x3, stride=2, feature maps=64, 
(3)~MaxPool3x3, stride=1, 
(4)~Conv3x3, stride=1, feature maps=32,  
(5)~Conv1x1, stride=1, feature maps=32,  
(6)~Conv1x1, stride=1, feature maps=2,  
(7)~Softmax. 

Our adversarial network is fully convolutional, and has been designed such that the receptive field of the last layer neurons in $R_{\boldsymbol \theta}$ and $D_{\boldsymbol \phi}$ are similar.
We first train the $R_{\boldsymbol \theta}$ network with just self-regularization loss for $1,000$ steps, and $D_{\boldsymbol \phi}$ for $200$ steps. 
Then, for each update of $D_{\boldsymbol \phi}$, we update $R_{\boldsymbol \theta}$ twice, \ie $K_d$ is set to $1$, and $K_g$ is set to $50$ in Algorithm~\ref{alg:training_GAN_w_L1}.

The eye gaze estimation network is similar to~\cite{Zhang15a}, with some changes to enable it to better exploit our large synthetic dataset.
The input is a $35 \times 55$ grayscale image that is passed through 5 convolutional layers followed by 3 fully connected layers,  the last one encoding the 3-dimensional gaze vector:
(1)~Conv3x3, feature maps=32, 
(2)~Conv3x3, feature maps=32, 
(3)~Conv3x3, feature maps=64, 
(4)~MaxPool3x3, stride=2, 
(5)~Conv3x3, feature maps=80,  
(6)~Conv3x3, feature maps=192,  
(7)~MaxPool2x2, stride=2,  
(8)~FC9600,  
(9)~FC1000, 
(10)~FC3, 
(11)~Euclidean loss.
All networks are trained with a constant $0.001$ learning rate and $512$ batch size, until the validation error converges.

\subsection{Hand Pose Estimation from Depth Images}
Next, we evaluate our method for hand pose estimation in depth images. 
We use the NYU hand pose dataset~\cite{tompson14NYU} that contains $72,757$ training frames and $8,251$ testing frames captured by $3$ Kinect cameras -- one frontal and 2 side views. 
Each depth frame is labeled with hand pose information that has been used to create a synthetic depth image. 
We pre-process the data by cropping the pixels from real images using the synthetic images. 
The images are resized to $224 \times 224$ before passing them to the ConvNet.

\vspace{-0.15in}
\paragraph{Qualitative Results:}
Figure~\ref{fig:results_qualitative_hand} shows example output of SimGAN on the NYU hand pose test set. 
The main source of noise in real depth images is from depth discontinuity at the edges, which the SimGAN is able to learn without requiring any label information.

\begin{figure}
\centering
\vspace{-0.2cm}
\includegraphics[width=1.1\linewidth]{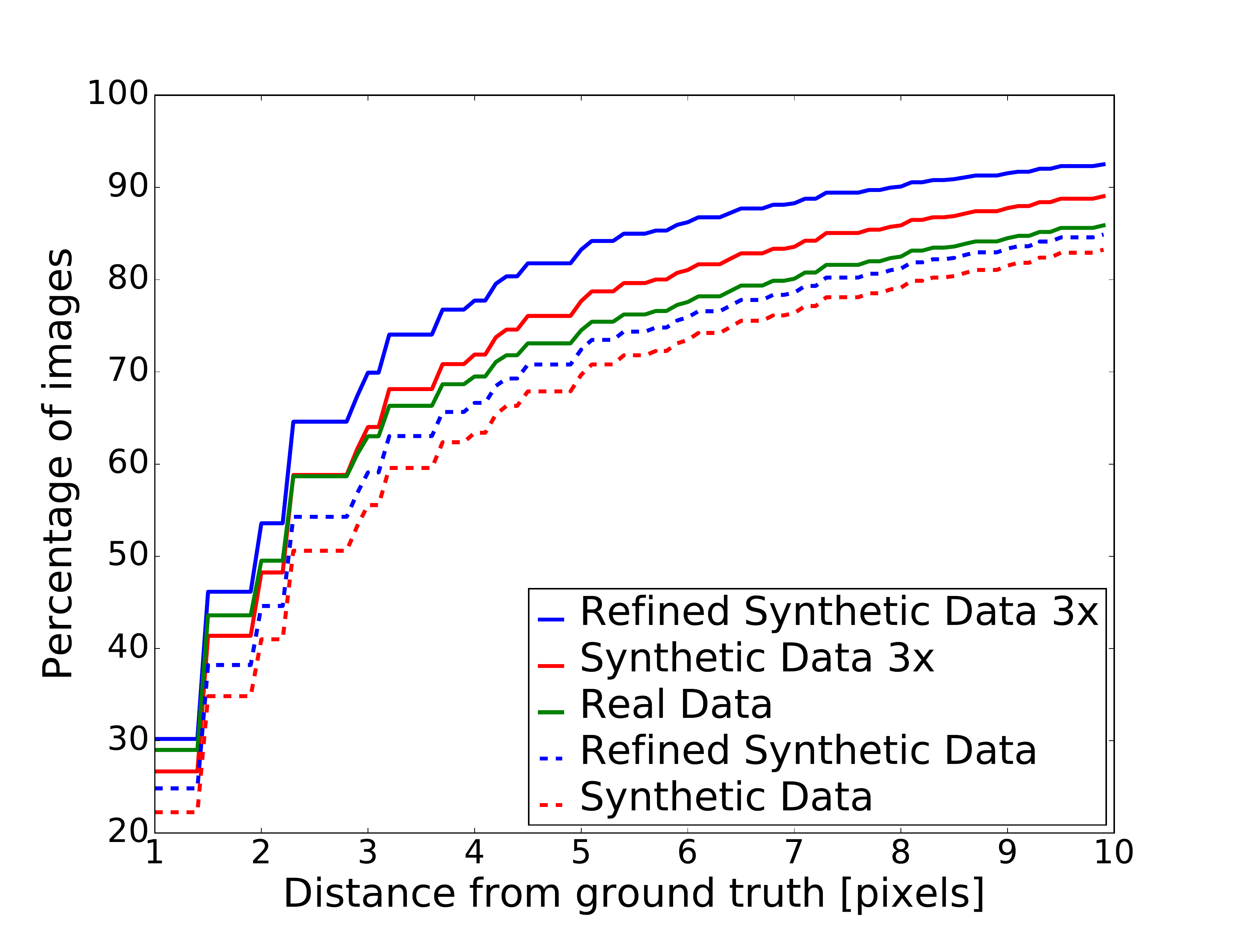}
\vspace{-0.6cm}
\caption{Quantitative results for hand pose estimation on the NYU hand pose test set of real depth images~\cite{tompson14NYU}.
The plot shows cumulative curves as a function of distance from ground truth keypoint locations, for different numbers of training examples of synthetic and refined images.
}
\label{fig:handpose_result_quantitative}
\end{figure}

\vspace{-0.15in}
\paragraph{Quantitative Results:} 
We train a fully convolutional hand pose estimator CNN similar to Stacked Hourglass Net~\cite{Newell16} on real, synthetic and refined synthetic images of the NYU hand pose training set, and evaluate each model on all \emph{real} images in the NYU hand pose test set.
We train on the same 14 hand joints as in~\cite{tompson14NYU}.
Many state-of-the-art hand pose estimation methods are customized pipelines that consist of several steps.  
We use only a single deep neural network to analyze the effect of improving the synthetic images to avoid bias due to other factors. 
Figure~\ref{fig:handpose_result_quantitative} and Table~\ref{tab:handpose_result_table} present quantitative results on NYU hand pose.
Training on refined synthetic data -- the output of SimGAN which does not require any labeling for the real images --  outperforms the model trained on real images with supervision, by $8.8 \%$.
The proposed method also outperforms training on synthetic data.
We also observe a large improvement as the number of synthetic training examples is increased -- here 3x corresponds to training on all views.

\begin{table}
\centering
\begin{tabular}{|l|c|} \hline
Training data  & \% of images within $d$  \\ \hline 
Synthetic Data & 69.7 \\
Refined Synthetic Data & 72.4 \\
Real Data & 74.5 \\
Synthetic Data 3x & 77.7 \\
Refined Synthetic Data 3x & {\bf 83.3} \\
\hline
\end{tabular}
\caption{Comparison of a hand pose estimator trained on synthetic data, real data, and the output of SimGAN.
The results are at distance $d=5$ pixels from ground truth.
}
\label{tab:handpose_result_table}
\end{table}

\vspace{-0.1in}
\paragraph{Implementation Details:}
The architecture is the same as for eye gaze estimation, except the input image size is $224 \times 224$, filter size is $7 \times 7$, and $10$ ResNet blocks are used.
The discriminative net $D_{\boldsymbol \phi}$ is:
(1)~Conv7x7, stride=4, feature maps=96, 
(2)~Conv5x5, stride=2, feature maps=64, 
(3)~MaxPool3x3, stride=2, 
(4)~Conv3x3, stride=2, feature maps=32,  
(5)~Conv1x1, stride=1, feature maps=32,  
(6)~Conv1x1, stride=1, feature maps=2,  
(7)~Softmax. 
We train the $R_{\boldsymbol \theta}$ network first with just  self-regularization loss for $500$ steps and $D_{\boldsymbol \phi}$ for $200$ steps; then, for each update of $D_{\boldsymbol \phi}$ we update $R_{\boldsymbol \theta}$ twice, \ie $K_d$ is set to $1$, and $K_g$ is set to $2$ in Algorithm~\ref{alg:training_GAN_w_L1}.

For hand pose estimation, we use the Stacked Hourglass Net of~\cite{Newell16} $2$ hourglass blocks, and an output heatmap size $64 \times 64$. 
We augment at training time with random $[-20,20]$ degree rotations and crops. 

\subsection{Ablation Study}

\begin{figure}
\centering
\includegraphics[width=1.0\linewidth]{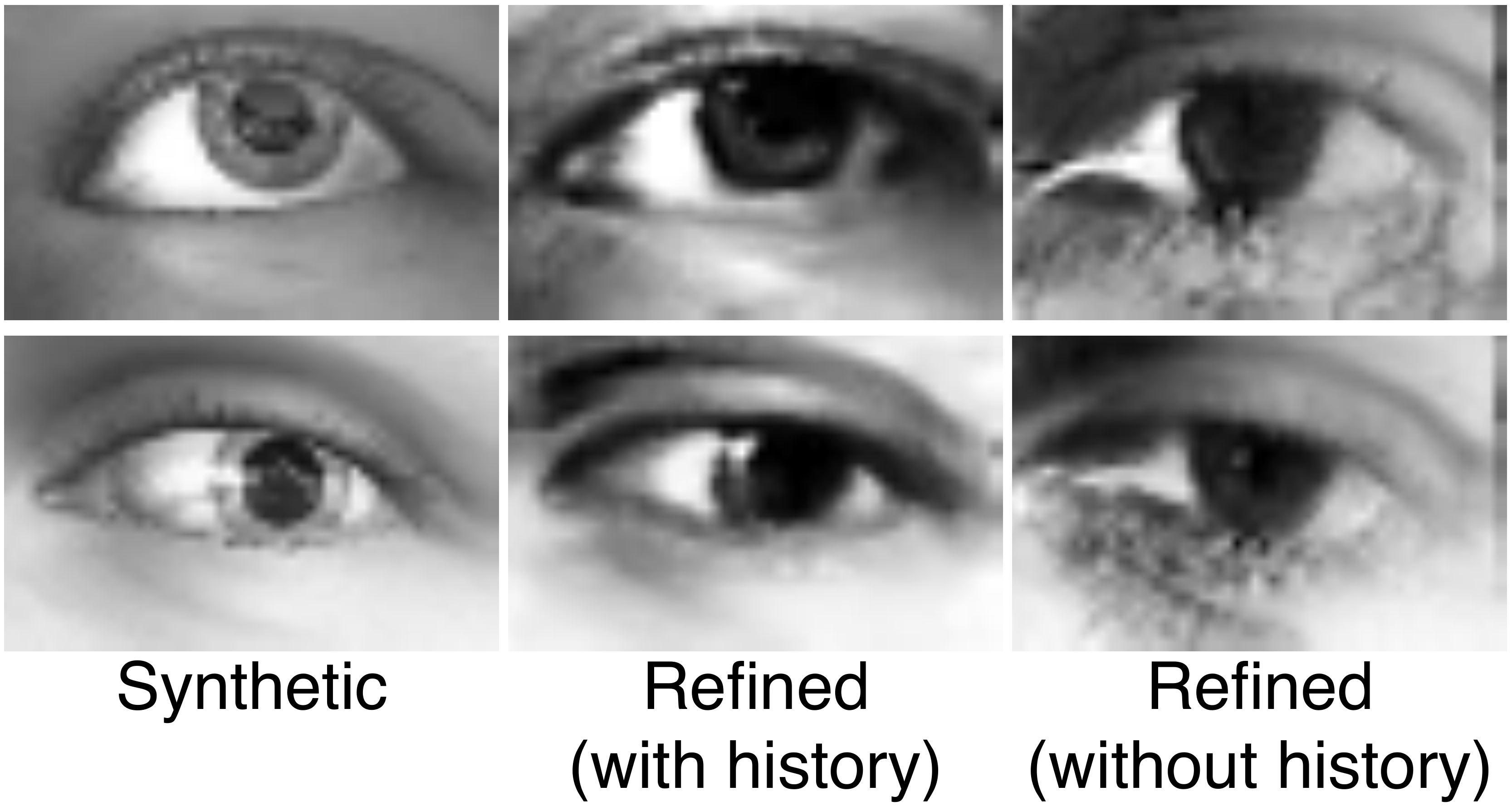} \\
\caption{Using a history of refined images for updating the discriminator. 
(Left) synthetic images; (middle) result of using the history of refined images; (right) result without using a history of refined images (instead using only the most recent refined images). 
We observe obvious unrealistic artifacts, especially around the corners of the eyes.
}
\label{fig:history_vs_noHistory}
\end{figure}

First, we analyzed the effect of using history of refined images during training. As shown in Figure~\ref{fig:history_vs_noHistory}, using the history of refined images (second column) prevents severe artifacts observed while training without the history (third column). 
This results in an increased gaze estimation error of $12.2$ degrees without the history, in comparison to $7.8$ degrees with the history.

\begin{figure}
\centering

\includegraphics[width=0.8\linewidth]{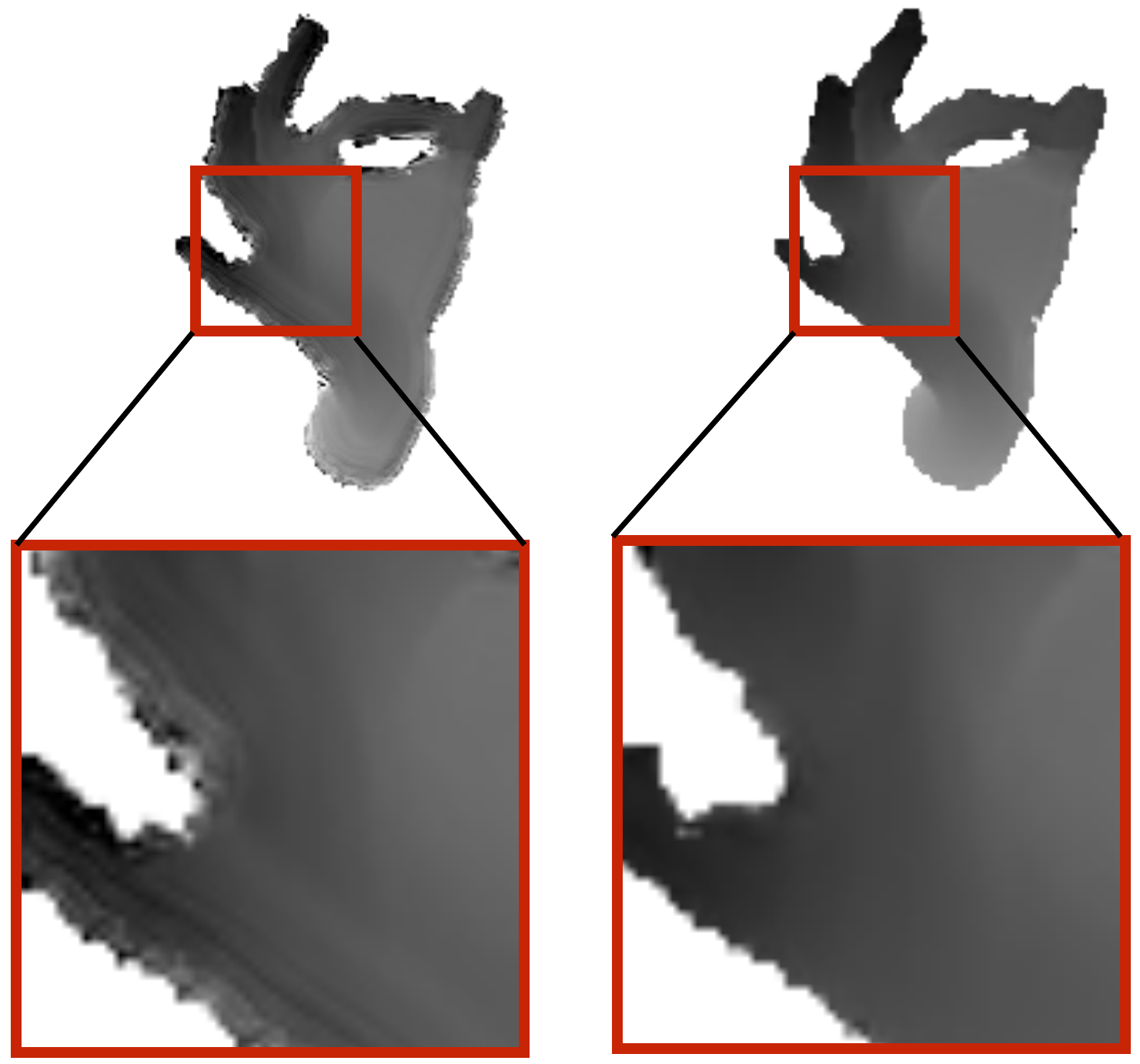} \\
\vskip-0.05in
Global adversarial loss \hskip0.1in Local adversarial loss 
\vskip1pt
\caption{Importance of using a local adversarial loss. 
(Left) an example image that has been generated with a standard `global' adversarial loss on the whole image. 
The noise around the edge of the hand contains obvious unrealistic depth boundary artifacts. 
(Right) the same image generated with a local adversarial loss that looks significantly more realistic. 
}
\label{fig:global_vs_local}
\end{figure}

Next, we compare local vs global adversarial loss during training. A global adversarial loss uses a fully connected layer in the discriminator network, classifying the whole image as real vs refined. The local adversarial loss removes the artifacts and makes the generated image significantly more realistic, as seen in Figure~\ref{fig:global_vs_local}.

\section{Conclusions and Future Work}
We have proposed Simulated+Unsupervised learning to add realism to the simulator while preserving the annotations of the synthetic images.
We described SimGAN, our method for S+U learning, that uses an adversarial network and demonstrated state-of-the-art results without any labeled real data. 
In future, we intend to explore modeling the noise distribution to generate more than one refined image for each synthetic image, and investigate refining videos rather than single images. 

\paragraph{Acknowledgement:} 
We are grateful to our colleagues Barry Theobald, Carlos Guestrin, Ruslan Salakhutdinov, Abhishek Sharma and Yin Zhou for their valuable inputs.

\pagebreak
\bibliographystyle{ieee}
\bibliography{mybib}

\begin{thebibliography}{10}\itemsep=-1pt

\bibitem{youtub8M_16}
S.~Abu{-}El{-}Haija, N.~Kothari, J.~Lee, P.~Natsev, G.~Toderici,
  B.~Varadarajan, and S.~Vijayanarasimhan.
\newblock Youtube-8m: {A} large-scale video classification benchmark.
\newblock {\em arXiv preprint arXiv:1609.08675}, 2016.

\bibitem{Chen16}
X.~Chen, Y.~Duan, R.~Houthooft, J.~Schulman, I.~Sutskever, and P.~Abbeel.
\newblock {InfoGAN}: Interpretable representation learning by information
  maximizing generative adversarial nets.
\newblock {\em arXiv preprint arXiv:1606.03657}, 2016.

\bibitem{Darrell15}
T.~Darrell, P.~Viola, and G.~Shakhnarovich.
\newblock Fast pose estimation with parameter sensitive hashing.
\newblock In {\em Proc. CVPR}, 2015.

\bibitem{imagenet_cvpr09}
J.~Deng, W.~Dong, R.~Socher, L.-J. Li, K.~Li, and L.~Fei-Fei.
\newblock {ImageNet: A Large-Scale Hierarchical Image Database}.
\newblock In {\em Proc. CVPR}, 2009.

\bibitem{Gaidon16}
A.~Gaidon, Q.~Wang, Y.~Cabon, and E.~Vig.
\newblock Virtual worlds as proxy for multi-object tracking analysis.
\newblock In {\em Proc. CVPR}, 2016.

\bibitem{Ganin14}
Y.~Ganin and V.~Lempitsky.
\newblock Unsupervised domain adaptation by backpropagation.
\newblock {\em arXiv preprint arXiv:1409.7495}, 2014.

\bibitem{Gatys16}
L.~Gatys, A.~Ecker, and M.~Bethge.
\newblock Image style transfer using convolutional neural networks.
\newblock In {\em Proc. CVPR}, 2016.

\bibitem{Goodfellow14}
I.~Goodfellow, J.~Pouget-Abadie, M.~Mirza, B.~Xu, D.~Warde-Farley, S.~Ozair,
  A.~Courville, and Y.~Bengio.
\newblock Generative adversarial nets.
\newblock In {\em Proc. NIPS}, 2014.

\bibitem{Gupta16}
A.~Gupta, A.~Vedaldi, and A.~Zisserman.
\newblock Synthetic data for text localisation in natural images.
\newblock {\em Proc. CVPR}, 2016.

\bibitem{Gupta14}
S.~Gupta, R.~Girshick, P.~Arbel{\'a}ez, and J.~Malik.
\newblock Learning rich features from rgb-d images for object detection and
  segmentation.
\newblock In {\em Proc. ECCV}, 2014.

\bibitem{Handa16}
A.~Handa, V.~Patraucean, V.~Badrinarayanan, S.~Stent, and R.~Cipolla.
\newblock {SceneNet}: Understanding real world indoor scenes with synthetic
  data.
\newblock In {\em Proc. CVPR}, 2015.

\bibitem{He2015}
K.~He, X.~Zhang, S.~Ren, and J.~Sun.
\newblock Deep residual learning for image recognition.
\newblock {\em arXiv preprint arXiv:1512.03385}, 2015.

\bibitem{Im2015}
D.~J. Im, C.~D. Kim, H.~Jiang, and R.~Memisevic.
\newblock Generating images with recurrent adversarial networks.
\newblock {\em http://arxiv.org/abs/ 1602.05110}, 2016.

\bibitem{Ionescu14}
C.~Ionescu, D.~Papava, V.~Olaru, and C.~Sminchisescu.
\newblock Human3.6m: Large scale datasets and predictive methods for 3d human
  sensing in natural environments.
\newblock {\em PAMI}, 36(7):1325--1339, 2014.

\bibitem{Jaderberg16}
M.~Jaderberg, K.~Simonyan, A.~Vedaldi, and A.~Zisserman.
\newblock Reading text in the wild with convolutional neural networks.
\newblock {\em IJCV}, 116(1):1--20, 2016.

\bibitem{johnson11cg2real}
M.~K. Johnson, K.~Dale, S.~Avidan, H.~Pfister, W.~T. Freeman, and W.~Matusik.
\newblock Cg2real: Improving the realism of computer generated images using a
  large collection of photographs.
\newblock {\em {IEEE} Transactions on Visualization and Computer Graphics},
  17(9):1273--1285, 2011.

\bibitem{openimages}
I.~Krasin, T.~Duerig, N.~Alldrin, A.~Veit, S.~Abu-El-Haija, S.~Belongie,
  D.~Cai, Z.~Feng, V.~Ferrari, V.~Gomes, A.~Gupta, D.~Narayanan, C.~Sun,
  G.~Chechik, and K.~Murphy.
\newblock {OpenImages}: A public dataset for large-scale multi-label and
  multi-class image classification.
\newblock {\em Dataset available from https://github.com/openimages}, 2016.

\bibitem{LeCun04a}
Y.~LeCun, F.~Huang, and L.~Bottou.
\newblock Learning methods for generic object recognition with invariance to
  pose and lighting.
\newblock In {\em Proc. CVPR}, 2004.

\bibitem{Li2016}
C.~Li and M.~Wand.
\newblock Precomputed real-time texture synthesis with markovian generative
  adversarial networks.
\newblock In {\em Proc. ECCV}, 2016.

\bibitem{mscoco}
T.-Y. Lin, M.~Maire, S.~Belongie, J.~Hays, P.~Perona, D.~Ramanan,
  P.~Doll{\'a}r, and C.~L. Zitnick.
\newblock Microsoft {COCO}: Common objects in context.
\newblock In {\em Proc. ECCV}, 2014.

\bibitem{liu2016coupled}
M.-Y. Liu and O.~Tuzel.
\newblock Coupled generative adversarial networks.
\newblock In {\em Proc. NIPS}, 2016.

\bibitem{Lotter15}
W.~Lotter, G.~Kreiman, and D.~Cox.
\newblock Unsupervised learning of visual structure using predictive generative
  networks.
\newblock {\em arXiv preprint arXiv:1511.06380}, 2015.

\bibitem{Lu2014}
F.~Lu, Y.~Sugano, T.~Okabe, and Y.~Sato.
\newblock Adaptive linear regression for appearance-based gaze estimation.
\newblock {\em PAMI}, 36(10):2033--2046, 2014.

\bibitem{Nagaraja2016}
V.~K. Nagaraja, V.~I. Morariu, and L.~S. Davis.
\newblock Modeling context between objects for referring expression
  understanding.
\newblock In {\em Proc. ECCV}, 2016.

\bibitem{Newell16}
A.~Newell, K.~Yang, and J.~Deng.
\newblock Stacked hourglass networks for human pose estimation.
\newblock {\em arXiv preprint arXiv:1603.06937}, 2016.

\bibitem{Park15}
D.~Park and D.~Ramanan.
\newblock Articulated pose estimation with tiny synthetic videos.
\newblock In {\em Proc. CVPR}, 2015.

\bibitem{Peng15}
X.~Peng, B.~Sun, K.~Ali, and K.~Saenko.
\newblock Learning deep object detectors from 3d models.
\newblock In {\em Proc. ICCV}, 2015.

\bibitem{Pishchulin12}
L.~Pishchulin, A.~Jain, M.~Andriluka, T.~Thorm\"ahlen, and B.~Schiele.
\newblock Articulated people detection and pose estimation: Reshaping the
  future.
\newblock In {\em Proc. CVPR}, 2012.

\bibitem{qiu2016unrealcv}
W.~Qiu and A.~Yuille.
\newblock {UnrealCV}: Connecting computer vision to {Unreal Engine}.
\newblock {\em arXiv preprint arXiv:1609.01326}, 2016.

\bibitem{Rogez16}
G.~Rogez and C.~Schmid.
\newblock {MoCap}-guided data augmentation for 3d pose estimation in the wild.
\newblock {\em arXiv preprint arXiv:1607.02046}, 2016.

\bibitem{Ros_2016_CVPR}
G.~Ros, L.~Sellart, J.~Materzynska, D.~Vazquez, and A.~M. Lopez.
\newblock The {SYNTHIA Dataset}: A large collection of synthetic images for
  semantic segmentation of urban scenes.
\newblock In {\em Proc. CVPR}, 2016.

\bibitem{Salimans16}
T.~Salimans, I.~Goodfellow, W.~Zaremba, V.~Cheung, A.~Radford, and X.~Chen.
\newblock Improved techniques for training gans.
\newblock {\em arXiv preprint arXiv:1606.03498}, 2016.

\bibitem{Schneider2014}
T.~Schneider, B.~Schauerte, and R.~Stiefelhagen.
\newblock Manifold alignment for person independent appearance-based gaze
  estimation.
\newblock In {\em Proc. ICPR}, 2014.

\bibitem{Shafaei16}
A.~Shafaei, J.~Little, and M.~Schmidt.
\newblock Play and learn: Using video games to train computer vision models.
\newblock In {\em Proc. BMVC}, 2016.

\bibitem{Shotton13}
J.~Shotton, R.~Girshick, A.~Fitzgibbon, T.~Sharp, M.~Cook, M.~Finocchio,
  R.~Moore, P.~Kohli, A.~Criminisi, A.~Kipman, and A.~Blake.
\newblock Efficient human pose estimation from single depth images.
\newblock {\em PAMI}, 35(12):2821--2840, 2013.

\bibitem{Sugano2014}
Y.~Sugano, Y.~Matsushita, and Y.~Sato.
\newblock Learning-by-synthesis for appearance-based 3d gaze estimation.
\newblock In {\em Proc. CVPR}, 2014.

\bibitem{Supancic15}
J.~Supancic, G.~Rogez, Y.~Yang, J.~Shotton, and D.~Ramanan.
\newblock Depth-based hand pose estimation: data, methods, and challenges.
\newblock In {\em Proc. CVPR}, 2015.

\bibitem{tompson14NYU}
J.~Tompson, M.~Stein, Y.~Lecun, and K.~Perlin.
\newblock Real-time continuous pose recovery of human hands using convolutional
  networks.
\newblock {\em ACM Trans. Graphics}, 2014.

\bibitem{Tuzel16}
O.~Tuzel, Y.~Taguchi, and J.~Hershey.
\newblock Global-local face upsampling network.
\newblock {\em arXiv preprint arXiv:1603.07235}, 2016.

\bibitem{vandenOord16}
A.~van~den Oord, N.~Kalchbrenner, and K.~Kavukcuoglu.
\newblock Pixel recurrent neural networks.
\newblock {\em arXiv preprint arXiv:1601.06759}, 2016.

\bibitem{Wang2016}
X.~Wang and A.~Gupta.
\newblock Generative image modeling using style and structure adversarial
  networks.
\newblock In {\em Proc. ECCV}, 2016.

\bibitem{Wang15}
Z.~Wang, J.~Yang, H.~Jin, E.~Shechtman, A.~Agarwala, J.~Brandt, and T.~Huang.
\newblock Deepfont: Identify your font from an image.
\newblock In {\em Proc. ACMM}, 2015.

\bibitem{Wood16}
E.~Wood, T.~Baltru{\v{s}}aitis, L.~Morency, P.~Robinson, and A.~Bulling.
\newblock Learning an appearance-based gaze estimator from one million
  synthesised images.
\newblock In {\em Proc. ACM Symposium on Eye Tracking Research \&
  Applications}, 2016.

\bibitem{Yoo16}
D.~Yoo, N.~Kim, S.~Park, A.~Paek, and I.~Kweon.
\newblock Pixel-level domain transfer.
\newblock In {\em Proc. ECCV}, 2016.

\bibitem{SeqGan_Yu16}
L.~Yu, W.~Zhang, J.~Wang, and Y.~Yu.
\newblock Seqgan: Sequence generative adversarial nets with policy gradient.
\newblock {\em arXiv preprint arXiv:1609.05473}, 2016.

\bibitem{Zhang15}
X.~Zhang, Y.~Fu, A.~Zang, L.~Sigal, and G.~Agam.
\newblock Learning classifiers from synthetic data using a multichannel
  autoencoder.
\newblock {\em arXiv preprint arXiv:1503.03163}, 2015.

\bibitem{Zhang15a}
X.~Zhang, Y.~Sugano, M.~Fritz, and A.~Bulling.
\newblock Appearance-based gaze estimation in the wild.
\newblock In {\em Proc. CVPR}, 2015.

\bibitem{ZhangLL16}
Y.~Zhang, K.~Lee, and H.~Lee.
\newblock Augmenting supervised neural networks with unsupervised objectives
  for large-scale image classification.
\newblock In {\em Proc. ICML}, 2016.

\bibitem{zhu2016generative}
J.-Y. Zhu, P.~Kr{\"a}henb{\"u}hl, E.~Shechtman, and A.~Efros.
\newblock Generative visual manipulation on the natural image manifold.
\newblock In {\em Proc. ECCV}, 2016.

\end{thebibliography}


\clearpage

\section*{Additional Experiments }

\subsection*{Qualitative Experiments for Appearance-based Gaze Estimation}

\begin{figure*}
\centering
\newcommand\expimagewidth{0.115}
\begin{tabular}{cccccccc}

\includegraphics[width=\expimagewidth\linewidth]{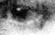} & \hspace{-0.1em}\includegraphics[width=\expimagewidth\linewidth]{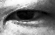} & \hspace{-0.1em}\includegraphics[width=\expimagewidth\linewidth]{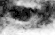} & \hspace{-0.1em}\includegraphics[width=\expimagewidth\linewidth]{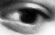} & \hspace{-0.1em}\includegraphics[width=\expimagewidth\linewidth]{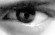} & \hspace{-0.1em}\includegraphics[width=\expimagewidth\linewidth]{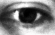} & \hspace{-0.1em}\includegraphics[width=\expimagewidth\linewidth]{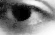} \vspace{0.01cm} \\
\includegraphics[width=\expimagewidth\linewidth]{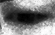} & \hspace{-0.1em}\includegraphics[width=\expimagewidth\linewidth]{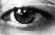} & \hspace{-0.1em}\includegraphics[width=\expimagewidth\linewidth]{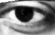} & \hspace{-0.1em}\includegraphics[width=\expimagewidth\linewidth]{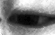} & \hspace{-0.1em}\includegraphics[width=\expimagewidth\linewidth]{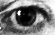} & \hspace{-0.1em}\includegraphics[width=\expimagewidth\linewidth]{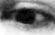} & \hspace{-0.1em}\includegraphics[width=\expimagewidth\linewidth]{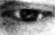} \vspace{0.01cm} \\
\includegraphics[width=\expimagewidth\linewidth]{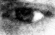} & \hspace{-0.1em}\includegraphics[width=\expimagewidth\linewidth]{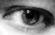} & \hspace{-0.1em}\includegraphics[width=\expimagewidth\linewidth]{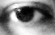} & \hspace{-0.1em}\includegraphics[width=\expimagewidth\linewidth]{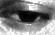} & \hspace{-0.1em}\includegraphics[width=\expimagewidth\linewidth]{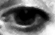} & \hspace{-0.1em}\includegraphics[width=\expimagewidth\linewidth]{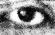} & \hspace{-0.1em}\includegraphics[width=\expimagewidth\linewidth]{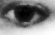} \vspace{0.01cm} \\
\includegraphics[width=\expimagewidth\linewidth]{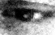} & \hspace{-0.1em}\includegraphics[width=\expimagewidth\linewidth]{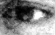} & \hspace{-0.1em}\includegraphics[width=\expimagewidth\linewidth]{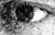} & \hspace{-0.1em}\includegraphics[width=\expimagewidth\linewidth]{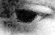} & \hspace{-0.1em}\includegraphics[width=\expimagewidth\linewidth]{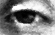} & \hspace{-0.1em}\includegraphics[width=\expimagewidth\linewidth]{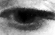} & \hspace{-0.1em}\includegraphics[width=\expimagewidth\linewidth]{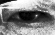} \vspace{0.01cm} \\
\end{tabular}
\caption{Example real images from MPIIGaze dataset.}
\label{fig:real_MPII_gaze}
\end{figure*}

\paragraph{Dataset:} The gaze estimation dataset consists of 1.2M synthetic images from eye gaze synthesizer UnityEyes~\cite{Wood16} and 214K real images from the MPIIGaze dataset~\cite{Zhang15a} -- samples shown in Figure~\ref{fig:real_MPII_gaze}.
MPIIGaze is a very challenging eye gaze estimation dataset captured under extreme illumination conditions.
For UnityEyes we use a single generic rendering environment to generate training data without any dataset-specific targeting.

\begin{figure*}
\centering
\newcommand\expimagewidth{0.115}
\begin{tabular}{cccccccc}

\rotatebox{90}{\small Synthetic}  \hspace{0.01em} \includegraphics[width=\expimagewidth\linewidth]{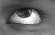} & \hspace{-0.1em}\includegraphics[width=\expimagewidth\linewidth]{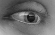} & \hspace{-0.1em}\includegraphics[width=\expimagewidth\linewidth]{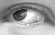} & \hspace{-0.1em}\includegraphics[width=\expimagewidth\linewidth]{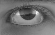} & \hspace{-0.1em}\includegraphics[width=\expimagewidth\linewidth]{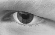} & \hspace{-0.1em}\includegraphics[width=\expimagewidth\linewidth]{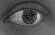} & \hspace{-0.1em}\includegraphics[width=\expimagewidth\linewidth]{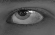} \vspace{0.01cm} \\
\rotatebox{90}{\small Refined} \hspace{0.2em} \includegraphics[width=\expimagewidth\linewidth]{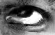} & \hspace{-0.1em}\includegraphics[width=\expimagewidth\linewidth]{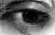} & \hspace{-0.1em}\includegraphics[width=\expimagewidth\linewidth]{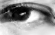} & \hspace{-0.1em}\includegraphics[width=\expimagewidth\linewidth]{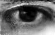} & \hspace{-0.1em}\includegraphics[width=\expimagewidth\linewidth]{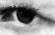} & \hspace{-0.1em}\includegraphics[width=\expimagewidth\linewidth]{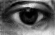} & \hspace{-0.1em}\includegraphics[width=\expimagewidth\linewidth]{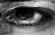} \vspace{0.2cm} \\
\hline  \\
\rotatebox{90}{\small Synthetic}  \hspace{0.01em} \includegraphics[width=\expimagewidth\linewidth]{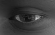} & \hspace{-0.1em}\includegraphics[width=\expimagewidth\linewidth]{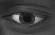} & \hspace{-0.1em}\includegraphics[width=\expimagewidth\linewidth]{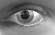} & \hspace{-0.1em}\includegraphics[width=\expimagewidth\linewidth]{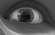} & \hspace{-0.1em}\includegraphics[width=\expimagewidth\linewidth]{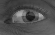} & \hspace{-0.1em}\includegraphics[width=\expimagewidth\linewidth]{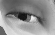} & \hspace{-0.1em}\includegraphics[width=\expimagewidth\linewidth]{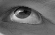} \vspace{0.01cm} \\
\rotatebox{90}{\small Refined} \hspace{0.2em} \includegraphics[width=\expimagewidth\linewidth]{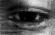} & \hspace{-0.1em}\includegraphics[width=\expimagewidth\linewidth]{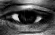} & \hspace{-0.1em}\includegraphics[width=\expimagewidth\linewidth]{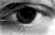} & \hspace{-0.1em}\includegraphics[width=\expimagewidth\linewidth]{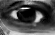} & \hspace{-0.1em}\includegraphics[width=\expimagewidth\linewidth]{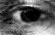} & \hspace{-0.1em}\includegraphics[width=\expimagewidth\linewidth]{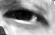} & \hspace{-0.1em}\includegraphics[width=\expimagewidth\linewidth]{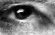} \vspace{0.2cm} \\
\hline  \\
\rotatebox{90}{\small Synthetic}  \hspace{0.01em} \includegraphics[width=\expimagewidth\linewidth]{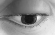} & \hspace{-0.1em}\includegraphics[width=\expimagewidth\linewidth]{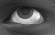} & \hspace{-0.1em}\includegraphics[width=\expimagewidth\linewidth]{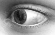} & \hspace{-0.1em}\includegraphics[width=\expimagewidth\linewidth]{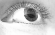} & \hspace{-0.1em}\includegraphics[width=\expimagewidth\linewidth]{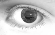} & \hspace{-0.1em}\includegraphics[width=\expimagewidth\linewidth]{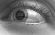} & \hspace{-0.1em}\includegraphics[width=\expimagewidth\linewidth]{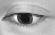} \vspace{0.01cm} \\
\rotatebox{90}{\small Refined} \hspace{0.2em} \includegraphics[width=\expimagewidth\linewidth]{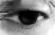} & \hspace{-0.1em}\includegraphics[width=\expimagewidth\linewidth]{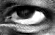} & \hspace{-0.1em}\includegraphics[width=\expimagewidth\linewidth]{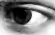} & \hspace{-0.1em}\includegraphics[width=\expimagewidth\linewidth]{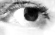} & \hspace{-0.1em}\includegraphics[width=\expimagewidth\linewidth]{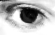} & \hspace{-0.1em}\includegraphics[width=\expimagewidth\linewidth]{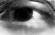} & \hspace{-0.1em}\includegraphics[width=\expimagewidth\linewidth]{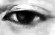} \vspace{0.2cm} \\
\hline  \\
\rotatebox{90}{\small Synthetic}  \hspace{0.01em} \includegraphics[width=\expimagewidth\linewidth]{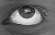} & \hspace{-0.1em}\includegraphics[width=\expimagewidth\linewidth]{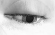} & \hspace{-0.1em}\includegraphics[width=\expimagewidth\linewidth]{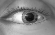} & \hspace{-0.1em}\includegraphics[width=\expimagewidth\linewidth]{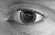} & \hspace{-0.1em}\includegraphics[width=\expimagewidth\linewidth]{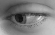} & \hspace{-0.1em}\includegraphics[width=\expimagewidth\linewidth]{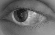} & \hspace{-0.1em}\includegraphics[width=\expimagewidth\linewidth]{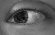} \vspace{0.01cm} \\
\rotatebox{90}{\small Refined} \hspace{0.2em} \includegraphics[width=\expimagewidth\linewidth]{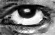} & \hspace{-0.1em}\includegraphics[width=\expimagewidth\linewidth]{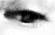} & \hspace{-0.1em}\includegraphics[width=\expimagewidth\linewidth]{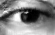} & \hspace{-0.1em}\includegraphics[width=\expimagewidth\linewidth]{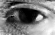} & \hspace{-0.1em}\includegraphics[width=\expimagewidth\linewidth]{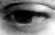} & \hspace{-0.1em}\includegraphics[width=\expimagewidth\linewidth]{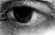} & \hspace{-0.1em}\includegraphics[width=\expimagewidth\linewidth]{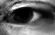} \vspace{0.2cm} \\
\hline  \\
\rotatebox{90}{\small Synthetic}  \hspace{0.01em} \includegraphics[width=\expimagewidth\linewidth]{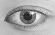} & \hspace{-0.1em}\includegraphics[width=\expimagewidth\linewidth]{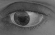} & \hspace{-0.1em}\includegraphics[width=\expimagewidth\linewidth]{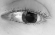} & \hspace{-0.1em}\includegraphics[width=\expimagewidth\linewidth]{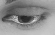} & \hspace{-0.1em}\includegraphics[width=\expimagewidth\linewidth]{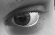} & \hspace{-0.1em}\includegraphics[width=\expimagewidth\linewidth]{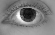} & \hspace{-0.1em}\includegraphics[width=\expimagewidth\linewidth]{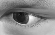} \vspace{0.01cm} \\
\rotatebox{90}{\small Refined} \hspace{0.2em} \includegraphics[width=\expimagewidth\linewidth]{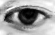} & \hspace{-0.1em}\includegraphics[width=\expimagewidth\linewidth]{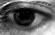} & \hspace{-0.1em}\includegraphics[width=\expimagewidth\linewidth]{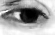} & \hspace{-0.1em}\includegraphics[width=\expimagewidth\linewidth]{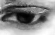} & \hspace{-0.1em}\includegraphics[width=\expimagewidth\linewidth]{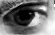} & \hspace{-0.1em}\includegraphics[width=\expimagewidth\linewidth]{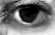} & \hspace{-0.1em}\includegraphics[width=\expimagewidth\linewidth]{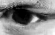} \vspace{0.2cm} \\
\hline  \\
\rotatebox{90}{\small Synthetic}  \hspace{0.01em} \includegraphics[width=\expimagewidth\linewidth]{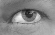} & \hspace{-0.1em}\includegraphics[width=\expimagewidth\linewidth]{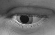} & \hspace{-0.1em}\includegraphics[width=\expimagewidth\linewidth]{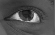} & \hspace{-0.1em}\includegraphics[width=\expimagewidth\linewidth]{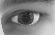} & \hspace{-0.1em}\includegraphics[width=\expimagewidth\linewidth]{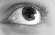} & \hspace{-0.1em}\includegraphics[width=\expimagewidth\linewidth]{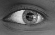} & \hspace{-0.1em}\includegraphics[width=\expimagewidth\linewidth]{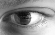} \vspace{0.01cm} \\
\rotatebox{90}{\small Refined} \hspace{0.2em} \includegraphics[width=\expimagewidth\linewidth]{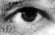} & \hspace{-0.1em}\includegraphics[width=\expimagewidth\linewidth]{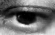} & \hspace{-0.1em}\includegraphics[width=\expimagewidth\linewidth]{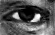} & \hspace{-0.1em}\includegraphics[width=\expimagewidth\linewidth]{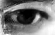} & \hspace{-0.1em}\includegraphics[width=\expimagewidth\linewidth]{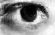} & \hspace{-0.1em}\includegraphics[width=\expimagewidth\linewidth]{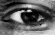} & \hspace{-0.1em}\includegraphics[width=\expimagewidth\linewidth]{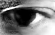} \vspace{0.2cm} \\
\hline  \\
\rotatebox{90}{\small Synthetic}  \hspace{0.01em} \includegraphics[width=\expimagewidth\linewidth]{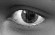} & \hspace{-0.1em}\includegraphics[width=\expimagewidth\linewidth]{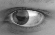} & \hspace{-0.1em}\includegraphics[width=\expimagewidth\linewidth]{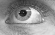} & \hspace{-0.1em}\includegraphics[width=\expimagewidth\linewidth]{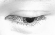} & \hspace{-0.1em}\includegraphics[width=\expimagewidth\linewidth]{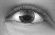} & \hspace{-0.1em}\includegraphics[width=\expimagewidth\linewidth]{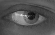} & \hspace{-0.1em}\includegraphics[width=\expimagewidth\linewidth]{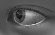} \vspace{0.01cm} \\
\rotatebox{90}{\small Refined} \hspace{0.2em} \includegraphics[width=\expimagewidth\linewidth]{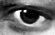} & \hspace{-0.1em}\includegraphics[width=\expimagewidth\linewidth]{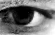} & \hspace{-0.1em}\includegraphics[width=\expimagewidth\linewidth]{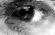} & \hspace{-0.1em}\includegraphics[width=\expimagewidth\linewidth]{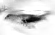} & \hspace{-0.1em}\includegraphics[width=\expimagewidth\linewidth]{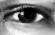} & \hspace{-0.1em}\includegraphics[width=\expimagewidth\linewidth]{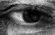} & \hspace{-0.1em}\includegraphics[width=\expimagewidth\linewidth]{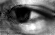} \vspace{0.2cm} \\
\end{tabular}
\caption{Qualitative results for automatic refinement of simulated eyes. The top row (in each set of two rows) shows the synthetic eye image, and the bottom row shows the corresponding refined image.}
\label{fig:more_results_qualitative_gaze}
\end{figure*}

\paragraph{Qualitative Results:}
In Figure~\ref{fig:more_results_qualitative_gaze}, we show many examples of synthetic, and refined images from the eye gaze dataset. We show many pairs of synthetic and refined in multiple rows. The top row contains synthetic images, and the bottom row contains corresponding refined images.
As shown, we observe a significant qualitative improvement of the synthetic images: SimGAN successfully captures the skin texture, sensor noise and the appearance of the iris region in the real images. Note that our method preserves the annotation information (gaze direction) while improving the realism.

\subsection*{Qualitative Experiments for Hand Pose Estimation}
\begin{figure*}
\centering
\newcommand\expimagewidth{0.14}
\begin{tabular}{cccccccc}
\includegraphics[width=\expimagewidth\linewidth]{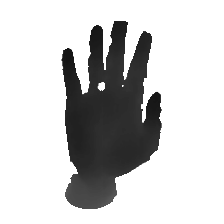} & \hspace{-0.5cm}\includegraphics[width=\expimagewidth\linewidth]{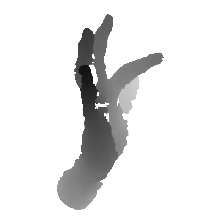} & \hspace{-0.5cm}\includegraphics[width=\expimagewidth\linewidth]{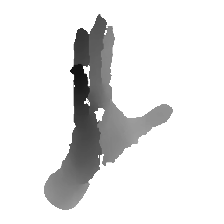} & \hspace{-0.5cm}\includegraphics[width=\expimagewidth\linewidth]{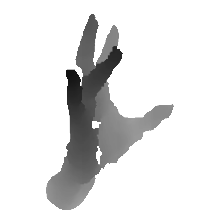} & \hspace{-0.5cm}\includegraphics[width=\expimagewidth\linewidth]{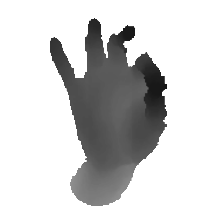} & \hspace{-0.5cm}\includegraphics[width=\expimagewidth\linewidth]{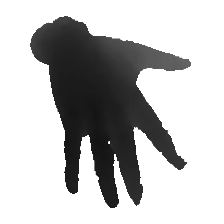} & \hspace{-0.5cm}\includegraphics[width=\expimagewidth\linewidth]{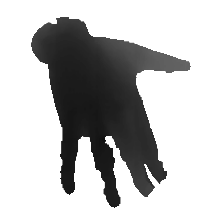} \vspace{0.01cm} \\
\includegraphics[width=\expimagewidth\linewidth]{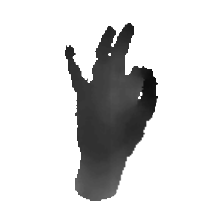} & \hspace{-0.5cm}\includegraphics[width=\expimagewidth\linewidth]{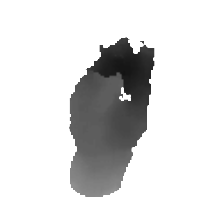} & \hspace{-0.5cm}\includegraphics[width=\expimagewidth\linewidth]{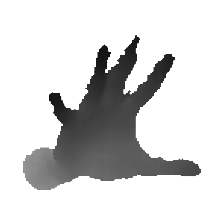} & \hspace{-0.5cm}\includegraphics[width=\expimagewidth\linewidth]{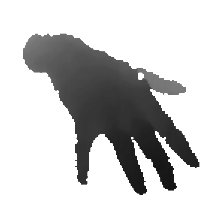} & \hspace{-0.5cm}\includegraphics[width=\expimagewidth\linewidth]{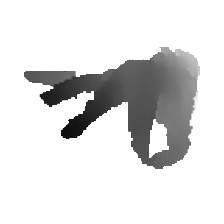} & \hspace{-0.5cm}\includegraphics[width=\expimagewidth\linewidth]{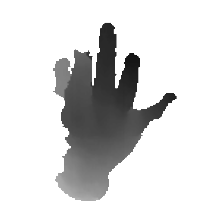} & \hspace{-0.5cm}\includegraphics[width=\expimagewidth\linewidth]{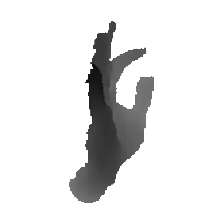} \vspace{0.01cm} \\
\includegraphics[width=\expimagewidth\linewidth]{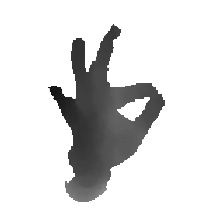} & \hspace{-0.5cm}\includegraphics[width=\expimagewidth\linewidth]{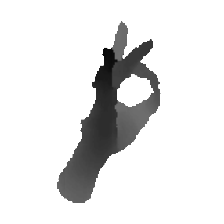} & \hspace{-0.5cm}\includegraphics[width=\expimagewidth\linewidth]{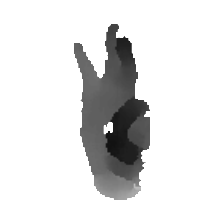} & \hspace{-0.5cm}\includegraphics[width=\expimagewidth\linewidth]{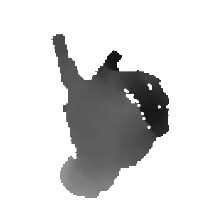} & \hspace{-0.5cm}\includegraphics[width=\expimagewidth\linewidth]{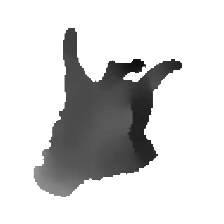} & \hspace{-0.5cm}\includegraphics[width=\expimagewidth\linewidth]{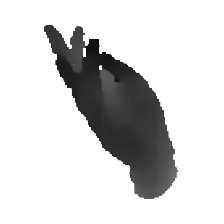} & \hspace{-0.5cm}\includegraphics[width=\expimagewidth\linewidth]{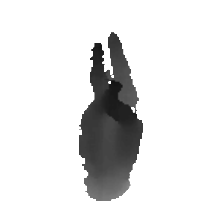} \vspace{0.01cm} \\
\end{tabular}
\caption{Example real test images in the NYU hand dataset.}
\label{fig:ex_real_nyu_hand}
\end{figure*}

\paragraph{Dataset:} Next, we evaluate our method for hand pose estimation in depth images. 
We use the NYU hand pose dataset~\cite{tompson14NYU} that contains $72,757$ training frames and $8,251$ testing frames.
Each depth frame is labeled with hand pose information that has been used to create a synthetic depth image. 
We pre-process the data by cropping the pixels from real images using the synthetic images. 
Figure~\ref{fig:ex_real_nyu_hand} shows example real depth images from the dataset. 
The images are resized to $224 \times 224$ before passing them to the refiner network.

\begin{figure*}
\centering
\newcommand\expimagewidth{0.14}
\begin{tabular}{cccccccc}
\rotatebox{90}{\small \;\;\;\;\;\; Synthetic}   \hspace{0.01em} \includegraphics[width=\expimagewidth\linewidth]{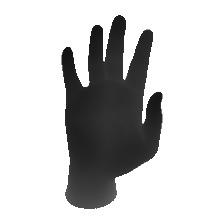} & \hspace{-0.5cm}\includegraphics[width=\expimagewidth\linewidth]{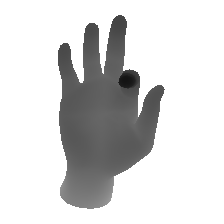} & \hspace{-0.5cm}\includegraphics[width=\expimagewidth\linewidth]{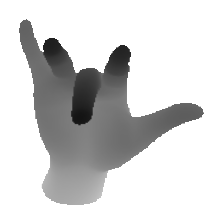} & \hspace{-0.5cm}\includegraphics[width=\expimagewidth\linewidth]{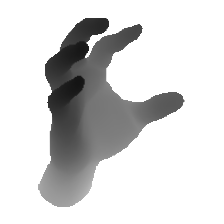} & \hspace{-0.5cm}\includegraphics[width=\expimagewidth\linewidth]{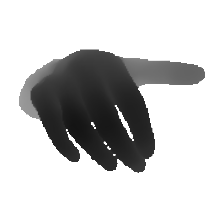} & \hspace{-0.5cm}\includegraphics[width=\expimagewidth\linewidth]{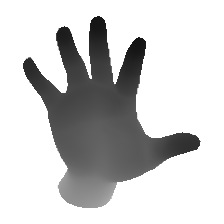} & \hspace{-0.5cm}\includegraphics[width=\expimagewidth\linewidth]{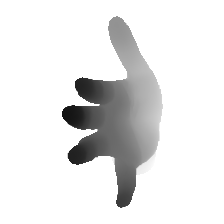} \vspace{0.01cm} \\
\rotatebox{90}{\small \;\;\;\;\;\;\;\; Refined} \hspace{0.2em} \includegraphics[width=\expimagewidth\linewidth]{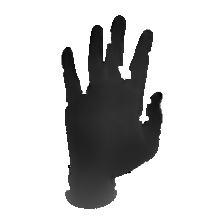} & \hspace{-0.5cm}\includegraphics[width=\expimagewidth\linewidth]{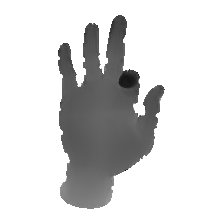} & \hspace{-0.5cm}\includegraphics[width=\expimagewidth\linewidth]{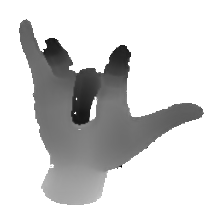} & \hspace{-0.5cm}\includegraphics[width=\expimagewidth\linewidth]{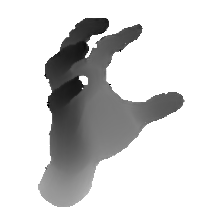} & \hspace{-0.5cm}\includegraphics[width=\expimagewidth\linewidth]{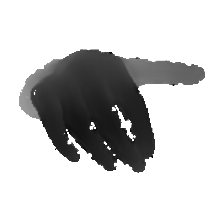} & \hspace{-0.5cm}\includegraphics[width=\expimagewidth\linewidth]{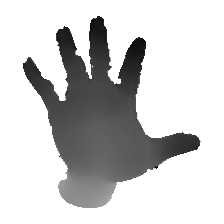} & \hspace{-0.5cm}\includegraphics[width=\expimagewidth\linewidth]{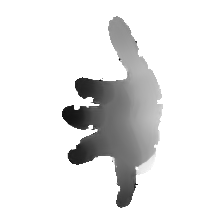} \vspace{0.2cm} \\
\hline  \\
\rotatebox{90}{\small \;\;\;\;\;\; Synthetic}   \hspace{0.01em} \includegraphics[width=\expimagewidth\linewidth]{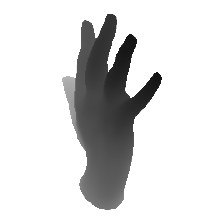} & \hspace{-0.5cm}\includegraphics[width=\expimagewidth\linewidth]{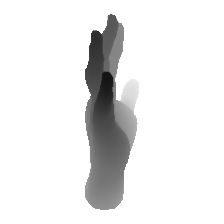} & \hspace{-0.5cm}\includegraphics[width=\expimagewidth\linewidth]{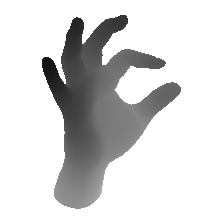} & \hspace{-0.5cm}\includegraphics[width=\expimagewidth\linewidth]{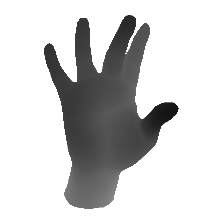} & \hspace{-0.5cm}\includegraphics[width=\expimagewidth\linewidth]{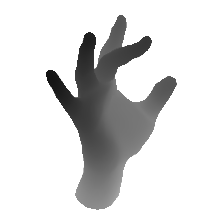} & \hspace{-0.5cm}\includegraphics[width=\expimagewidth\linewidth]{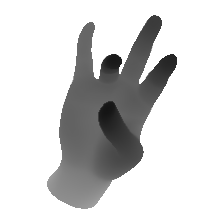} & \hspace{-0.5cm}\includegraphics[width=\expimagewidth\linewidth]{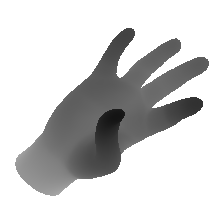} \vspace{0.01cm} \\
\rotatebox{90}{\small \;\;\;\;\;\;\;\; Refined} \hspace{0.2em} \includegraphics[width=\expimagewidth\linewidth]{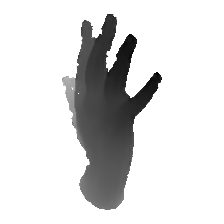} & \hspace{-0.5cm}\includegraphics[width=\expimagewidth\linewidth]{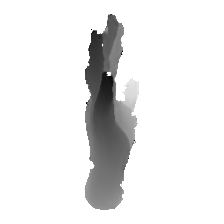} & \hspace{-0.5cm}\includegraphics[width=\expimagewidth\linewidth]{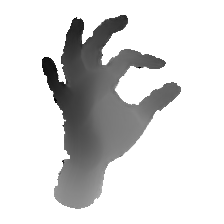} & \hspace{-0.5cm}\includegraphics[width=\expimagewidth\linewidth]{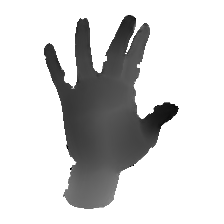} & \hspace{-0.5cm}\includegraphics[width=\expimagewidth\linewidth]{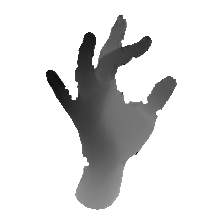} & \hspace{-0.5cm}\includegraphics[width=\expimagewidth\linewidth]{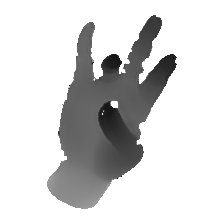} & \hspace{-0.5cm}\includegraphics[width=\expimagewidth\linewidth]{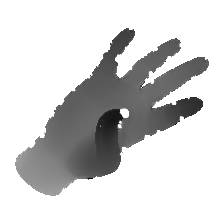} \vspace{0.2cm} \\
\hline  \\
\rotatebox{90}{\small \;\;\;\;\;\; Synthetic}   \hspace{0.01em} \includegraphics[width=\expimagewidth\linewidth]{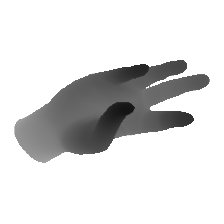} & \hspace{-0.5cm}\includegraphics[width=\expimagewidth\linewidth]{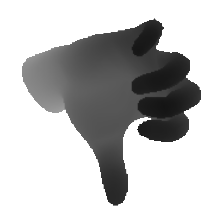} & \hspace{-0.5cm}\includegraphics[width=\expimagewidth\linewidth]{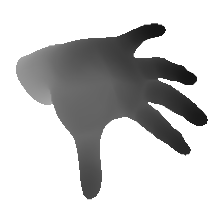} & \hspace{-0.5cm}\includegraphics[width=\expimagewidth\linewidth]{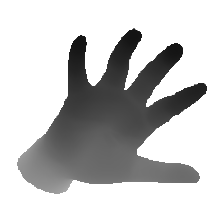} & \hspace{-0.5cm}\includegraphics[width=\expimagewidth\linewidth]{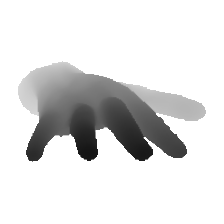} & \hspace{-0.5cm}\includegraphics[width=\expimagewidth\linewidth]{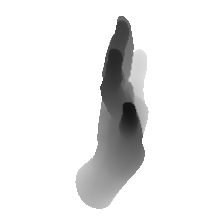} & \hspace{-0.5cm}\includegraphics[width=\expimagewidth\linewidth]{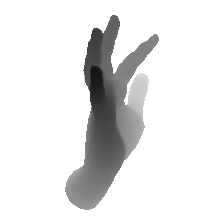} \vspace{0.01cm} \\
\rotatebox{90}{\small \;\;\;\;\;\;\;\; Refined} \hspace{0.2em} \includegraphics[width=\expimagewidth\linewidth]{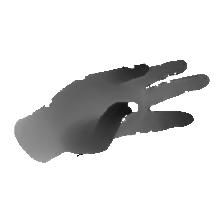} & \hspace{-0.5cm}\includegraphics[width=\expimagewidth\linewidth]{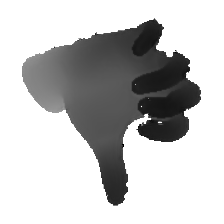} & \hspace{-0.5cm}\includegraphics[width=\expimagewidth\linewidth]{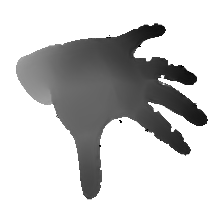} & \hspace{-0.5cm}\includegraphics[width=\expimagewidth\linewidth]{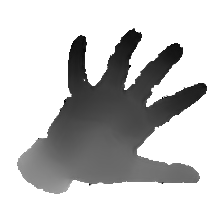} & \hspace{-0.5cm}\includegraphics[width=\expimagewidth\linewidth]{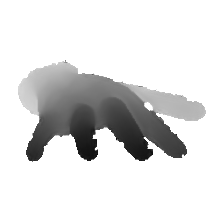} & \hspace{-0.5cm}\includegraphics[width=\expimagewidth\linewidth]{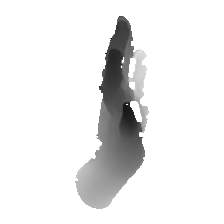} & \hspace{-0.5cm}\includegraphics[width=\expimagewidth\linewidth]{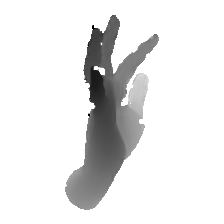} \vspace{0.2cm} \\
\hline  \\
\rotatebox{90}{\small \;\;\;\;\;\; Synthetic}   \hspace{0.01em} \includegraphics[width=\expimagewidth\linewidth]{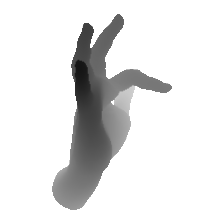} & \hspace{-0.5cm}\includegraphics[width=\expimagewidth\linewidth]{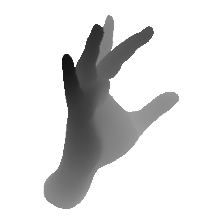} & \hspace{-0.5cm}\includegraphics[width=\expimagewidth\linewidth]{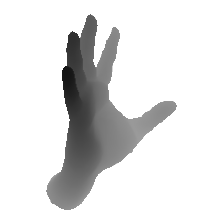} & \hspace{-0.5cm}\includegraphics[width=\expimagewidth\linewidth]{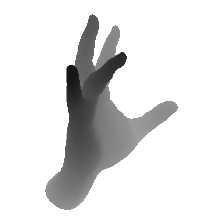} & \hspace{-0.5cm}\includegraphics[width=\expimagewidth\linewidth]{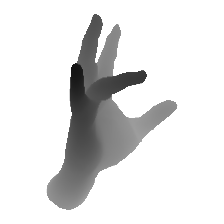} & \hspace{-0.5cm}\includegraphics[width=\expimagewidth\linewidth]{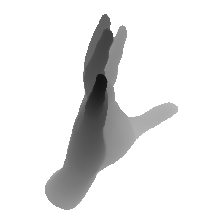} & \hspace{-0.5cm}\includegraphics[width=\expimagewidth\linewidth]{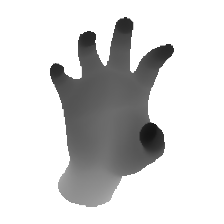} \vspace{0.01cm} \\
\rotatebox{90}{\small \;\;\;\;\;\;\;\; Refined} \hspace{0.2em} \includegraphics[width=\expimagewidth\linewidth]{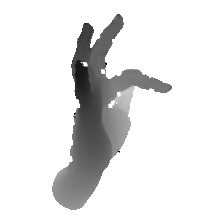} & \hspace{-0.5cm}\includegraphics[width=\expimagewidth\linewidth]{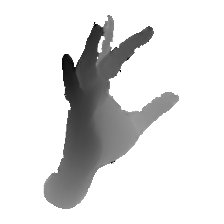} & \hspace{-0.5cm}\includegraphics[width=\expimagewidth\linewidth]{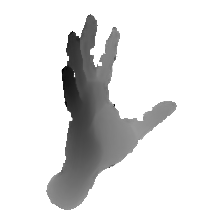} & \hspace{-0.5cm}\includegraphics[width=\expimagewidth\linewidth]{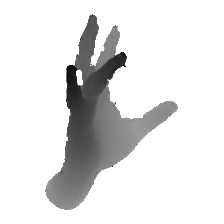} & \hspace{-0.5cm}\includegraphics[width=\expimagewidth\linewidth]{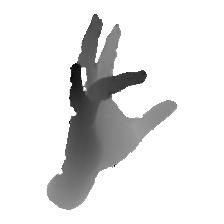} & \hspace{-0.5cm}\includegraphics[width=\expimagewidth\linewidth]{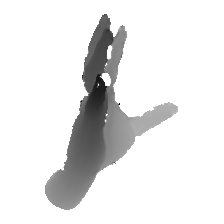} & \hspace{-0.5cm}\includegraphics[width=\expimagewidth\linewidth]{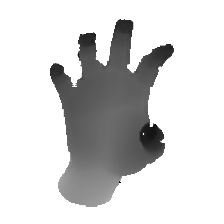} \vspace{0.2cm} \\
\hline  \\
\end{tabular}
\caption{Qualitative results for automatic refinement of NYU hand depth images. The top row (in each set of two rows) shows the synthetic hand image, and the bottom row is the corresponding refined image. Note how realistic the depth boundaries are compared to real images in Figure~\ref{fig:ex_real_nyu_hand}.}
\label{fig:more_results_qualitative_hand}
\end{figure*}

\paragraph{Quantative Results:} We show examples of synthetic and refined hand depth images in Figure~\ref{fig:more_results_qualitative_hand} from the test set. We show our results in multiple pairs of rows. The top row in each pair, contains synthetic depth image, and the bottom row shows the corresponding refined image using the proposed SimGAN approach. Note the realism added to the depth boundary in the refined images, compare to the real images in Figure~\ref{fig:ex_real_nyu_hand}.

\subsection*{Convergence Experiment}
\begin{figure*}
\centering
\includegraphics[width=\linewidth]{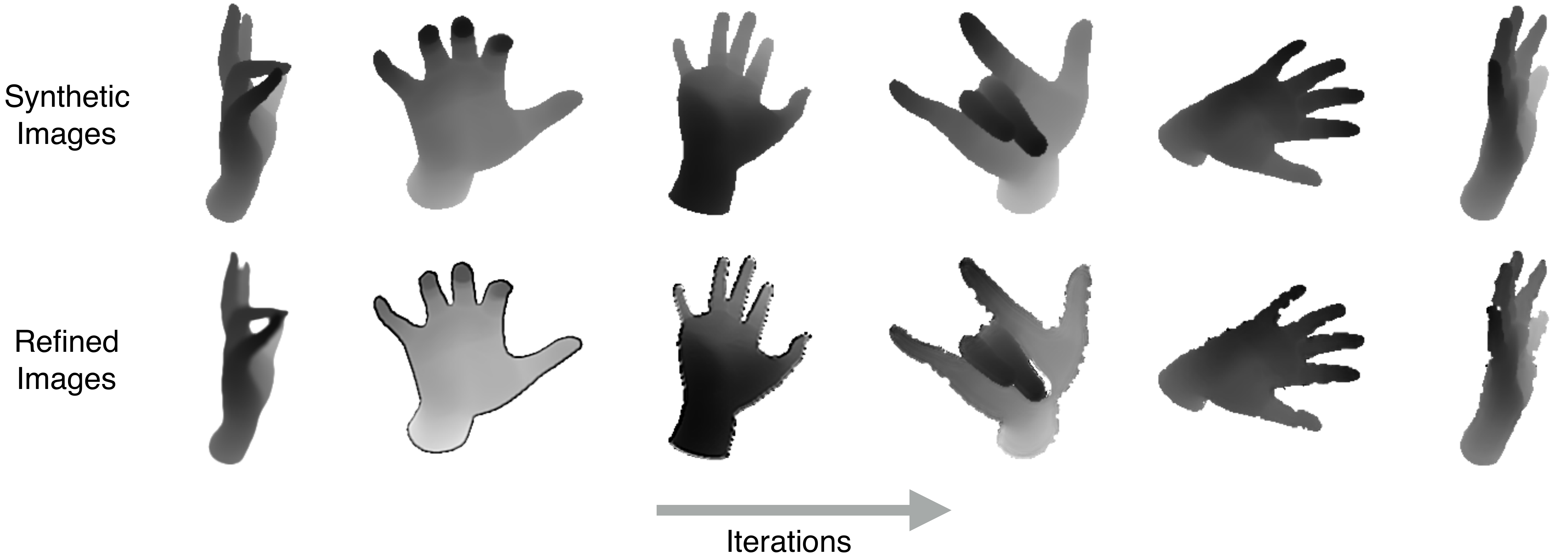} 
\caption{
SimGAN output as a function of training iterations for NYU hand pose.  
Columns correspond to increasing training iterations. 
First row shows synthetic images, and the second row shows corresponding refined images.
The first column is the result of training with $\ell_1$ image difference for $300$ steps; the later rows show the result when trained on top of this model. 
In the beginning the adversarial part of the cost introduces different kinds of unrealistic noise to try beat the adversarial network $D_{\boldsymbol \phi}$. 
As the dueling between $R_{\boldsymbol \theta}$ and $D_{\boldsymbol \phi}$ progresses, $R_{\boldsymbol \theta}$ learns to model the right kind of noise.}
\vspace{0.2cm}
\label{fig:results_iterations}
\end{figure*}

To investigate the convergence of our method, we visualize intermediate results as training progresses. 
As shown in Figure~\ref{fig:results_iterations}, in the beginning, the refiner network learns to predict very smooth edges using only the self-regularization loss. 
As the adversarial loss is enabled, the network starts adding artifacts at the depth boundaries. 
However, as these artifacts are not the same as real images, the discriminator easily learns to differentiate between the real and refined images. 
Slowly the network starts adding realistic noise, and after many steps, the refiner generates very realistic-looking images. 
We found it helpful to train the network with a low learning rate and for a large number of steps. 
For NYU hand pose we used lr=$0.0002$ in the beginning, and reduced to $0.00005$ after $600,000$ steps.

\end{document}